\documentclass{article}

\usepackage[eandd, final]{neurips_2026}

\usepackage[utf8]{inputenc} 
\usepackage[T1]{fontenc}    
\usepackage{hyperref}       
\usepackage{url}            
\usepackage{booktabs}       
\usepackage{bbm}            
\usepackage{nicefrac}       
\usepackage{microtype}      
\usepackage{xcolor}         
\usepackage[title]{appendix}
\usepackage[flushmargin]{footmisc}

\usepackage[table]{xcolor}   
\usepackage{multirow}
\usepackage{svg}

\definecolor{hA}{HTML}{F8F1ED} 
\definecolor{hB}{HTML}{F3D8CF} 
\definecolor{hC}{HTML}{E8AFA5} 
\definecolor{hD}{HTML}{CF7C78} 
\definecolor{hE}{HTML}{9F5A63} 
\definecolor{hF}{HTML}{5F3440} 

\newcommand{\hA}[1]{\cellcolor{hA}#1}
\newcommand{\hB}[1]{\cellcolor{hB}#1}
\newcommand{\hC}[1]{\cellcolor{hC}#1}
\newcommand{\hD}[1]{\cellcolor{hD}\textcolor{white}{#1}}
\newcommand{\hE}[1]{\cellcolor{hE}\textcolor{white}{#1}}
\newcommand{\hF}[1]{\cellcolor{hF}\textcolor{white}{#1}}

\definecolor{coolA}{HTML}{F2FBFC}
\definecolor{coolB}{HTML}{D8F0F3}
\definecolor{coolC}{HTML}{B5E1E8}
\definecolor{coolD}{HTML}{7FC8D6}
\definecolor{coolE}{HTML}{3AA6BD}
\definecolor{coolF}{HTML}{0B6F8A}

\newcommand{\oA}[1]{\cellcolor{coolA}{#1}}
\newcommand{\oB}[1]{\cellcolor{coolB}{#1}}
\newcommand{\oC}[1]{\cellcolor{coolC}{#1}}
\newcommand{\oD}[1]{\cellcolor{coolD}{#1}}
\newcommand{\oE}[1]{\cellcolor{coolE}{#1}}
\newcommand{\oF}[1]{\cellcolor{coolF}{\textcolor{white}{#1}}}

\usepackage{mathtools}
\usepackage{multirow}
\usepackage{graphicx}
\usepackage{amssymb}
\usepackage{amsfonts}
\usepackage{wrapfig}

\usepackage{pifont}
\usepackage{graphicx}
\usepackage{subcaption}
\newcommand{\cmark}{\ding{51}}%
\newcommand{\xmark}{\ding{55}}%
\newcommand{\drop}[1]{%
}

\newcommand{\otherdrop}[1]{%
    \textcolor{red}{\ensuremath{_{#1\%}}}%
}
 
\definecolor{CBBlue}{RGB}{0,114,178}
\definecolor{CBOrange}{RGB}{230,159,0}
\definecolor{CBGreen}{RGB}{0,158,115}
\definecolor{CBRed}{RGB}{213, 94, 0}
\definecolor{CBPink}{RGB}{204, 121, 167}
\definecolor{CBCyan}{RGB}{86, 180, 180}

\newcommand{\minisection}[1]{\vspace{0.0in} \noindent {\bf #1}}
\newcommand{\keyword}[1]{{\color{CBBlue}#1}}

\usepackage{pgfplots}
\pgfplotsset{compat=1.18}
\usetikzlibrary{patterns}

\definecolor{ibmBlue}{HTML}{648FFF}
\definecolor{ibmOrange}{HTML}{FE6100}
\usepackage{tcolorbox}
\tcbuselibrary{skins, breakable}
\usepackage{enumitem}

\title{GeoBiaset: A Counterfactual Benchmark for Demographic Bias in World-Level Geolocalization}

\author{%
Niccolò Niccoli$^{1}$ \quad
Silvia Dani$^{1}$ \quad
Marco Mistretta$^{1}$ 
\\
\textbf{Andrew D. Bagdanov$^{1}$ \quad
Marco Bertini$^{1}$ \quad
Lorenzo Seidenari$^{1}$}
\\[2mm]
$^{1}$University of Florence
}

\begin{document}

\maketitle

\begin{figure}[h]
    \vspace{-0.5cm}
    \centering
    \includegraphics[width=0.95\textwidth]{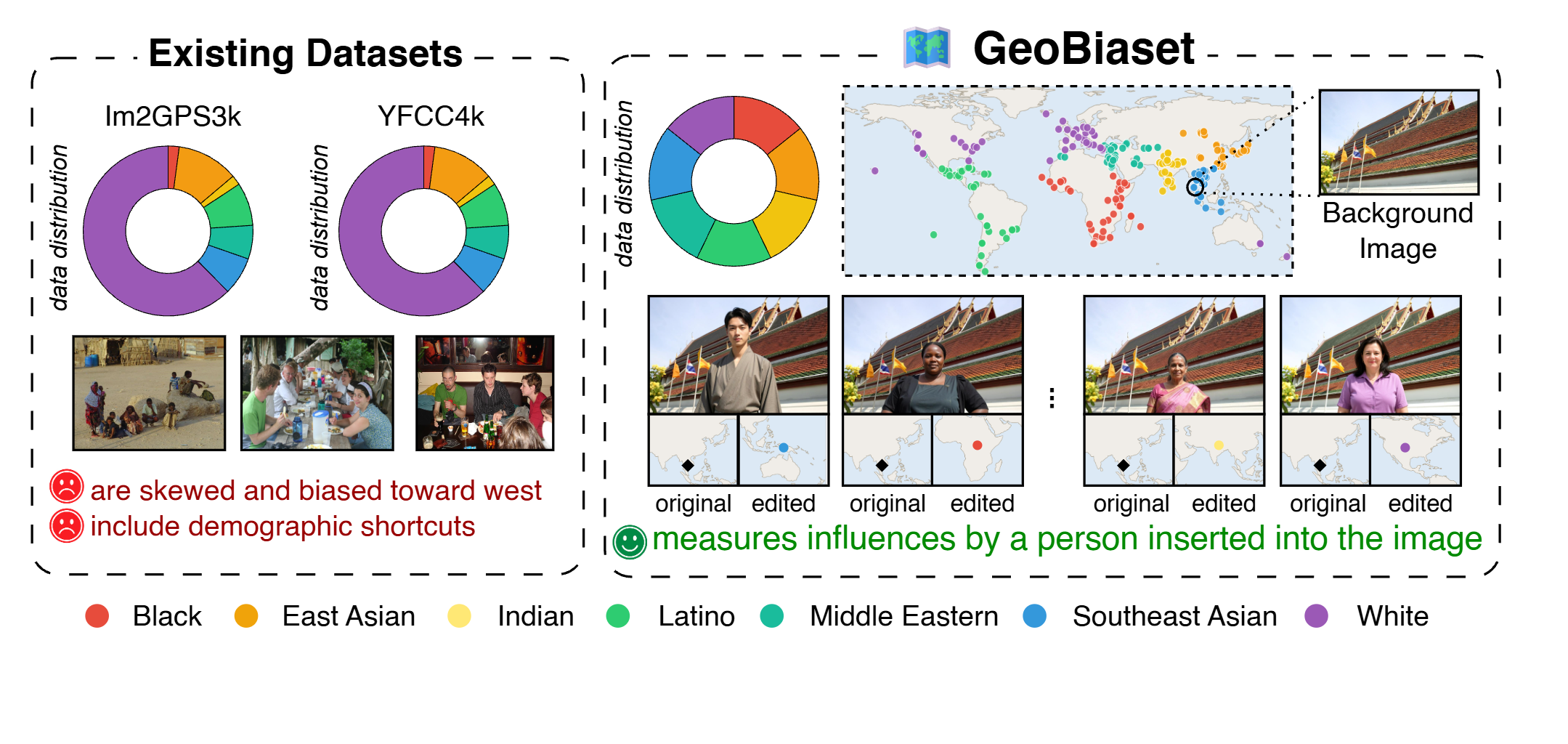}
    \vspace{-0.8cm}
    \caption{\textbf{Left:} Standard geolocalization benchmarks are imbalanced: scene cues and human attributes co-occur, making it unclear whether models rely on place-specific evidence or spurious demographic correlations.
\textbf{Right:} 
\textsc{GeoBiaset} generates counterfactual edits varying race, gender, and attire under fixed GPS coordinates. Black diamonds indicate predictions for original images; colored circles indicate predictions for edited images, with colors denoting the inserted person’s race category. Displacement toward race-associated regions highlights the audited bias.
    }
    \label{fig:teaser}
    \vspace{-0.1cm}
\end{figure}

\begin{abstract}
    World-level image geolocalization is an increasingly important task, yet the biases of current models remain poorly understood. Prior work has identified coarse dataset imbalances noting, for instance, that some cities are absent from training sets and that models tend to perform better in wealthier regions. However, no one has yet examined which visual cues drive these disparities. To do so, we introduce \textsc{GeoBiaset}, the first benchmark designed to measure how geolocalization models are influenced by the apparent race of a person inserted into the image foreground. The dataset consists of 9,541 images, pairing existing geotagged background images with counterfactual edits spanning seven races, holding the ground-truth coordinates fixed. Controlled masking experiments show that occlusion and background drift do not fully explain geolocalization degradation.
    Alongside the benchmark, we introduce the Person-Origin Perturbation Indicators (POPI), novel metrics that quantify model resilience to these demographic perturbations by measuring whether predictions are pulled toward the region associated with the inserted group. Our evaluation of state-of-the-art geolocalization and multimodal large language models reveals a 16-47\% drop in continent-level performance when a subject from a demographically mismatched group is inserted. A landmark versus no-landmark analysis and gradient-weighted heatmaps show that models preferentially attend to geographic landmarks when present, but shift attention to facial features in their absence.
\end{abstract}

\section{Introduction}
\label{sec:introduction}
Visual image geolocalization predicts where in the world a photograph was taken from its visual content alone. Its applications range from privacy auditing of social media imagery to disaster response and digital forensics. Building on the IM2GPS paradigm of matching a query against a large geotagged database~\cite{hays2008im2gps, vo2017revisiting}, modern systems combine contrastive vision-language encoders~\cite{vivanco2023geoclip}, retrieval-augmented generation~\cite{img2loc, jia2024g3}, and multimodal large language models~\cite{jia2025georanker, huang2025ai} to achieve strong accuracy on standard benchmarks.

Recent analyses demonstrate that planet-scale geolocalization models exhibit severe geographic bias, systematically favoring developed, Western, and urbanized regions over underrepresented ones~\cite{uribe2024regional, huang2025ai, li2025pixels, wang2025ai}. This results in substantial spatial disparities, including regional performance gaps of up to 42\%~\cite{qian2025earth}. While these systematic failures are widely attributed to uneven training data densities, prior work primarily quantifies macroscopic spatial distribution shifts without characterizing the specific visual mechanisms that drive individual prediction errors.

This is the gap we address. Faces and attire provide coarse geographic cues~\cite{haas2024pigeon, talreja2026georc}, but people travel, and apparent origin does not determine where a photo was taken. Standard benchmarks cannot reveal whether models rely on this signal, because geography, culture, clothing, and demographics co-occur in images, so correct predictions do not identify the driving cues.
To audit this sensitivity in world-level image geolocalization, we introduce \textsc{GeoBiaset}\footnote{
Dataset:~\url{https://kaggle.com/datasets/6393b4d11746fe3e4af72d5c71438ea2e003283fec6044064becdfd82a68475a}}, a counterfactual benchmark that varies inserted foreground subjects' demographics while holding ground-truth scene coordinates fixed.
Starting from 329 geotagged backgrounds from IM2GPS3k~\cite{vo2017revisiting} and YFCC4k~\cite{thomee2016yfcc100m}, we generate 28 edited variants by inserting a person from one of 7 mutually exclusive races, combined with 2 gender and 2 attire conditions (unconditioned and race-associated), for a total of 9541 image-annotation pairs. Faces are sampled from a demographically balanced source~\cite{karkkainen2019fairface} (see Fig.~\ref{fig:teaser}) and than the edited images pass automated screening for severe synthesis failures.

Additionally, we show that existing accuracy metrics are insensitive to the directional structure of these shifts. Therefore, we introduce the \textit{Person-Origin Perturbation Indicators} (POPI) to quantify the spatial distribution shift induced by inserting racially diverse subjects into the scene. We define three variants: \textbf{POPI-Attack (POPI-A)} measuring the rate at which an inserted subject shifts the prediction toward their associated geographic origin, indicating susceptibility to spurious visual correlations; \textbf{POPI-Help (POPI-H)} quantifying the frequency with which an inserted subject corrects an initially failed baseline prediction to the ground-truth region; and \textbf{POPI-Distraction (POPI-D)} evaluating how often an inserted subject misleads an initially correct prediction away from the ground truth.

We evaluate state-of-the-art geolocalization models (G3, GeoCLIP, Img2Loc) and a suite of open-weight MLLMs on \textsc{GeoBiaset}. We observe three consistent patterns. First, demographic foreground subjects act as strong distractors: inserting a mismatched-race subject reduces continent-level accuracy from between 16\% to 47\% relative to the original backgrounds, and the degradation persists when the background contains a recognized landmark. Second, for a subset of races, inserting a subject whose group matches the ground-truth region produces a measurable accuracy gain at the continent scale, indicating that models do not merely degrade under perturbation but actively use demographic cues as geographic proxies. Finally, gradient-weighted attention analysis reveals a stable pattern in which models attend to landmarks when present and shift reliance to the foreground face in their absence.

Our contributions are as follows:
\begin{itemize}
    \item We introduce \textsc{GeoBiaset}, a counterfactual benchmark for auditing demographic bias in worldwide image geolocalization via person insertions with fixed GPS labels.
    \item We propose POPI, a set of perturbation-based metrics that quantify whether demographic interventions attract, help, or distract geolocalization predictions relative to race-associated geographic regions.
    \item We evaluate seven geolocalization systems and show that foreground demographic cues can substantially reduce accuracy and redirect predictions.
    \item We analyze attire, face obfuscation, and landmark presence, showing that the effect is not limited to facial appearance and is amplified when scene-level geographic evidence is weak.
\end{itemize}
\section{Related Work}
\minisection{Image geolocalization.}
Image geolocalization infers where a photograph was taken from visual content alone. Early work matched queries against a large geotagged database~\cite{hays2008im2gps}; deep learning later reformulated the task as classification over discrete geographic cells~\cite{weyand2016planet, vo2017revisiting, seo2018cplanet} and revisited retrieval with stronger metric learning~\cite{berton2022deep, xu2023transvlad}. Current systems align images with coordinates in a joint embedding space~\cite{vivanco2023geoclip}, structure the prediction space through semantically grounded geocells~\cite{haas2024pigeon}, or layer retrieval-augmented and ranking-based reasoning on top of these representations~\cite{img2loc, jia2024g3, jia2025georanker}. Generalist MLLMs match specialized  models on standard benchmarks without task-specific training~\cite{huang2025ai, jay2025evaluating}. The field has focused on accuracy, robustness to viewpoint, and worldwide coverage.
Recently, several studies have begun to focus also on the privacy aspects of the task~\cite{liu2026geoshield, niccoli2026defending, wang2026hidingplainsightdiffusionbased, wu2026geocloak}.

\minisection{Bias in image geolocalization.} Social and demographic biases are well-known issues in artificial intelligence, especially from the advent of large multimodal models~\cite{huang2025visbias, kong2023mitigating, Zhang_2025_CVPR, gerych2024bendvlm, garcia2023uncurated, dehdashtian2024fairerclip, tanjim2024discovering}. A small but growing literature shows that geolocalization models do not err uniformly across the globe. Uribe et al.~\cite{uribe2024regional} document large accuracy drops on African imagery, with predictions shifted toward higher-income countries and the Global North. Huang et al.~\cite{huang2025ai} report that VLMs favor economically developed and densely populated regions and tend to over-predict well-known cities. IMAGEO-Bench~\cite{li2025pixels} confirms stronger performance in North America and Western Europe than in underrepresented areas, and KoreaGEO~\cite{wang2025ai} shows that even within a single country predictions collapse toward urban centers. EarthWhere~\cite{qian2025earth} measures gaps of up to 42\% between best and worst regions. These studies establish that performance is uneven across regions, countries, development level, and population density. They locate \textit{where} models fail. They do not characterize \textit{what visual content} drives prediction errors, which is the question \textsc{GeoBiaset} is built to answer.

\minisection{Counterfactual auditing of vision-language models.}
A complementary line of work audits VLMs through paired counterfactual images 
that vary a sensitive attribute while keeping the rest of the scene fixed. 
Earlier work used synthesized face images that vary race and gender while holding other visual factors fixed to measure demographic sensitivity in commercial classifiers~\cite{kim2021counterfactual}. 
SocialCounterfactuals~\cite{howard2024uncovering} synthesizes such pairs to 
isolate race and gender effects in VLM outputs, FOCUS~\cite{chen2026measuring} applies 
face-only edits over real photographs to study demographic disparities in 
occupational judgment, and Cultural Counterfactuals~\cite{howard2026cultural} 
varies cultural framing while holding identity constant.
Related synthetic counterfactual face benchmarks further study how
controlled demographic variation can support fairness evaluation while reducing confounding
\cite{ramesh2024synthetic}. These benchmarks 
quantify bias in decisions made \textit{about} the depicted person. We adopt 
the same experimental logic and apply it to a task in which the prediction 
target is a property of the \textit{place}, where the depicted person is in 
principle irrelevant. This shift in the target is what makes demographic 
influence in geolocalization a particularly subtle failure mode and what 
motivates the directional metrics introduced in Sec.~\ref{sec:protocol}.

\section{The GeoBiaset Dataset}
\label{sec:geobiaset}
Constructing \textsc{GeoBiaset} requires two sources of visual input: geographic background images and face images used to control the demographic appearance of the inserted subject. Background images are manually selected from two standard image geolocalization benchmarks, Im2GPS3k~\cite{hays2008im2gps} and YFCC4k~\cite{thomee2016yfcc100m}. These datasets are widely used in global image geolocalization, but they exhibit substantial geographic imbalance, with over-representation of Western regions~\cite{clark2023we, astruc2024openstreetview, uribe2024regional}. To reduce this source imbalance, we first partition the Earth into geographic areas and then sample an equal number of background images from each area after merging the two datasets. 

We further filter the selected images to remove cases in which a person is already present in the foreground, such as close-up portraits or images where human appearance could confound the demographic intervention. This filtering step is important because \textsc{GeoBiaset} aims to isolate the effect of the inserted subject rather than measure uncontrolled variation already present in the source image. After selection and filtering, the resulting background set contains 47 images for each of 7 geographic areas, for a total of 329 real images.
To define demographic categories in a reproducible manner, we adopt the racial taxonomy used by FairFace~\cite{karkkainen2019fairface}, which provides balanced annotations over seven categories: Black, East Asian, Indian, Middle Eastern, Latino, Southeast Asian, and White. We use FairFace because it offers an established, externally defined demographic categorization rather than relying on ad hoc labels introduced for this work. Our goal is not to treat these categories as biologically intrinsic, but to use them as socially salient visual attributes that may influence geolocation models.

\begin{figure}[h]
    \centering
   \includegraphics[width=0.95\linewidth]{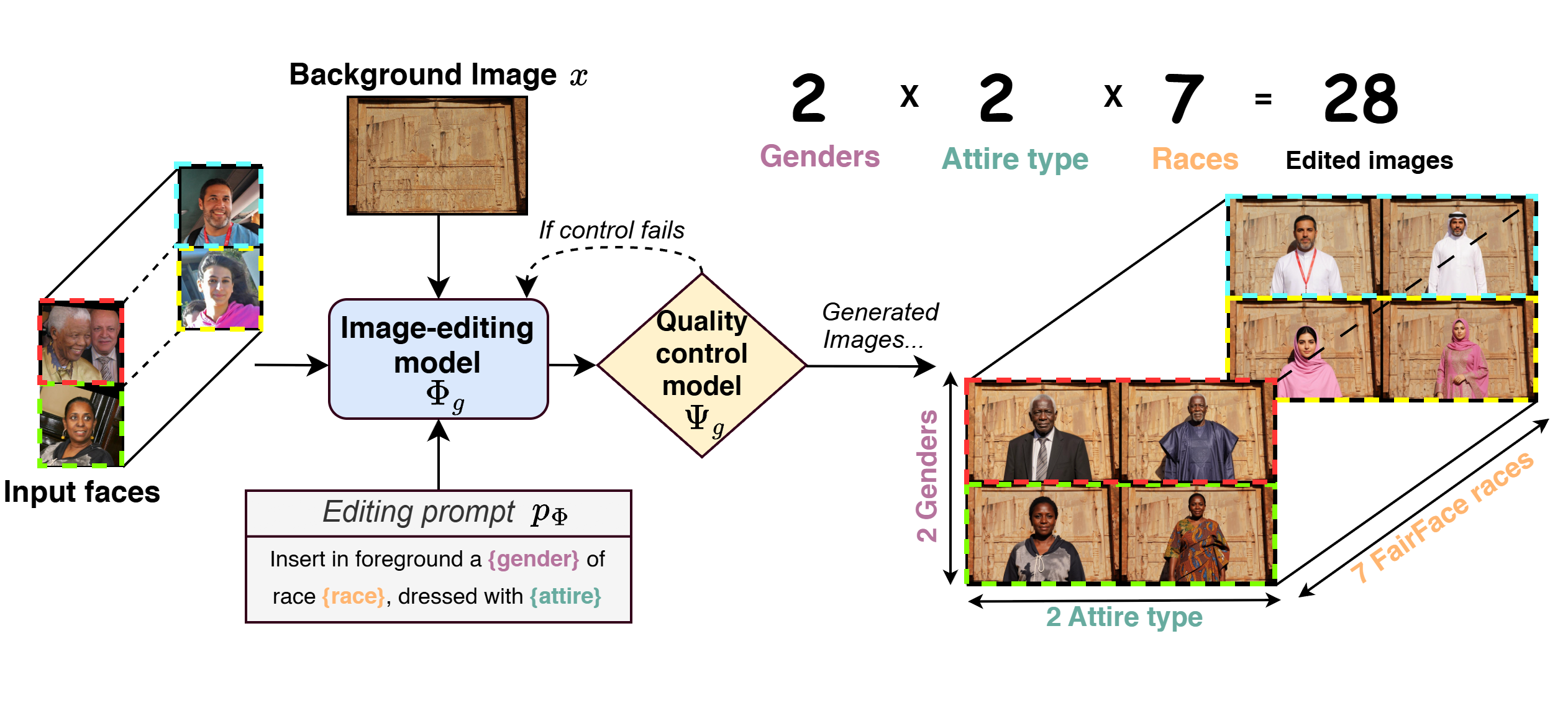}
    \vspace{-0.5cm}
    \caption{
    \textbf{Overview of the \textsc{GeoBiaset} image-generation pipeline.} Starting from a background image, an input face, and an editing prompt that specifies race, gender and optionally race-associated clothing we generate an edited image while preserving the overall geographic scene, without enforcing pixel-level accuracy. A multimodal quality-control module detects severe generation artifacts and triggers regeneration when needed. For each starting background image, we generate 28 edited images (2 genders × 2 attire types × 7 races).
    }
    \label{fig:generation_pipeline}
\end{figure}

We construct the face pool to isolate race as an independent variable while controlling for other visible attributes, including gender and age. We exclude identities below age 20, both to avoid using minors in bias-auditing interventions and to reduce age-related variation in facial morphology. We then group the remaining images by the variables race, gender, and age group, and manually select 10 high-quality images from each variable combination. This produces a sufficiently diverse pool from which face images can be sampled during generation while limiting excessive repetition of the same identity. By balancing the source face set across demographic groups, we reduce the risk that observed model behavior is driven by imbalance in the generation inputs rather than by demographic sensitivity in the geolocalization models under evaluation.

\minisection{Editing Pipeline.}
We design an image-editing pipeline that inserts human subjects into geographic scenes while preserving the visual evidence relevant to geolocalization, such as architecture, vegetation, road layout, terrain, lighting, and urban form. Let \(x\) denote a background image and let \(G_x=(\lambda_x,\phi_x)\) be its associated GPS coordinate, where \(\lambda_x\) and \(\phi_x\) denote longitude and latitude. Let \(r\) denote race, \(g\) denote gender, and \(f\) denote a face image sampled from the balanced face pool. We define the image-editing model as \(\Phi_g\) and the multimodal quality-control model as \(\Psi_g\). The complete pipeline forms a closed loop between synthesis and artifact detection, as shown in Fig.~\ref{fig:generation_pipeline}.

For each background image \(x\), demographic condition \((r,g)\), and sampled face image \(f\), we generate an edited image
\(
    y_{x,f} = \Phi_g(x, f, p_{\Phi}),
\)
where \(p_{\Phi}\) is a standardized editing prompt specifying the demographic attributes of the inserted person and instructing the model to preserve the original background, lighting, and photographic realism. Because the GPS label is inherited from the source background image, the edited image \(y_{x,f}\) should remain geolocalized at \(G_x\). This design allows us to measure whether changing the apparent demographic attributes of the person changes the predicted location while preserving the same underlying geographic scene as closely as possible. Residual background changes introduced by the editing process are quantified and discussed in Sec.~\ref{sec:causal_isolation}.

To study whether models rely on culturally or racially associated visual cues beyond facial appearance, we also construct a clothing-conditioned variant. For each race-gender pair \((r,g)\), we define a set of race-associated clothing examples \(C_{r,g}\). We then generate
\(
    y_{x,f,c} = \Phi_g(x, f, c, p_{\Phi}), \quad c \sim C_{r,g},
\)
where \(c\) is a sampled clothing description. This intervention enables us to evaluate whether geolocation models shift their predictions when the same geographic background is paired with clothing cues stereotypically associated with a particular demographic group.
In the attire-unconditioned setting, no clothing description is provided; however, the image-editing model may still generate race-associated attire through its learned demographic priors.

After each image is generated, we apply the multimodal quality-control model \(\Psi_g\) with prompt \(p_{\Psi}\) to detect severe synthesis failures, including missing body parts, duplicated limbs, multiple faces, poor blending, or cases where the inserted person is absent or too small to support the intended intervention. If the image is flagged as anomalous, the pipeline regenerates the image with the same conditioning variables but a different random seed, thereby sampling a different region of the generative model's latent space. To ensure termination, we allow up to 100 regeneration attempts.

Finally, we construct a face-obfuscated variant of \textsc{GeoBiaset} by applying Deface~\cite{orbhd_deface} to the generated images. This anonymized subset preserves the background scene, body pose, clothing, and overall image composition while blurring detectable faces, allowing us to test whether demographic perturbations persist when explicit facial appearance is removed.

\minisection{Implementation Details.}
We instantiate \(\Phi_g\) with FLUX.2-klein-9b-kv~\cite{flux-2-2025}, a conditioned diffusion model that accepts multiple reference images. The background \(x\) provides the geographic scene and the reference \(f\) controls the inserted subject's facial appearance. We use a standardized prompt specifying target race, gender, and age, conditioning on the reference face, and instructing the model to preserve background, lighting, pose naturalness, and realism. The prompt is reported in Appendix~\ref{prompt:image_editing}.

For clothing-conditioned images, we append a sampled attire description from a curated race--gender prompt set; the full attire lists and template are in Appendix~\ref{prompt:clothing}. These prompts are controlled visual interventions, not normative claims about demographic groups.

For \(\Psi_g\), we use Qwen3-VL-8B-Instruct~\cite{qwen3technicalreport} to detect synthesis failures listed above. Flagged images are regenerated with the same conditioning and a different seed. The prompt is in Appendix~\ref{prompt:quality_control}. This automatic filtering removes severe failures but is not a substitute for a human perceptual study and may miss subtle artifacts or culturally implausible insertions.

\minisection{Ex Post Quality Assessment.}\label{sec:quality}
This bring us to conduct an ex post quality assessment to verify that inserting human subjects does not introduce systematic degradation that could confound geolocalization performance. Artifacts such as blur, compression noise, poor blending, or hallucinated structures could alter visual cues independently of the intended demographic intervention.

We first evaluate generated images with CLIPIQA~\cite{wang2022exploring} and
QualiCLIP~\cite{agnolucci2024qualityaware}, two CLIP-based no-reference IQA metrics. Comparing edited images with their source background yields comparable quality scores, suggesting that interventions preserve photographic quality and do not introduce degradation that dominates downstream results. However, these no-reference metrics do not measure whether the geographic background itself is preserved.

We therefore additionally perform a full-reference comparison between each edited image and its corresponding source image, excluding the region occupied by the inserted subject. On the remaining background pixels, we obtain PSNR $20.17\pm3.11$ dB, SSIM $0.677\pm0.126$, and LPIPS $0.230\pm0.055$. These results reveal measurable pixel-level changes and show that the editing process does not guarantee strict background identity.
Importantly, the controlled masking experiment in Sec.~\ref{sec:causal_isolation} separately evaluates whether these residual background changes account for the observed geolocalization degradation.

See Appendix~\ref{app:full_grids} for full examples from
\textsc{GeoBiaset}. To support reproducibility, we have released our code.\footnote{Code:~\url{https://github.com/miccunifi/GeoBiaset}}

\section{Bias Evaluation Protocol}
\label{sec:protocol}
Our protocol compares predictions on original backgrounds and their edited variants in \textsc{GeoBiaset}. Since each edit inherits the GPS coordinate of its source, any prediction shift induced by the inserted person reflects sensitivity to the race intervention. Let \(x_i\) be a background with ground-truth GPS coordinate \(G_i=(\lambda_i,\phi_i)\). A geolocalization model \(\mathcal{M}\) predicts \(\hat{G}_i=\mathcal{M}(x_i)\), and we measure geographic error with the Haversine distance \(d(\cdot,\cdot)\).

For each \(x_i\), \textsc{GeoBiaset} provides 28 edited variants \(x_i^r=\Phi(x_i,r)\) in which a person from race group \(r\) is inserted while GPS label remains fixed and the underlying scene is preserved, though the generative pipeline does not guarantee strict pixel-level background identity. We use the seven FairFace categories: Black, East Asian, Indian, Middle Eastern, Latino, Southeast Asian, White~\cite{karkkainen2019fairface}. Race and gender follow FairFace terminology: race denotes a visually perceived attribute for controlled bias auditing, not a biological or self-identified label, while gender is a binary operational annotation for balancing and generation.

To measure whether predictions move in a systematic direction, each race group \(r\) is associated with a geographic support region \(R_r\), defined as the union of countries listed in Appendix~\ref{app:taxonomy}. These regions are evaluation constructs and should not be interpreted as intrinsic mappings between race and geography. For a prediction \(\hat{G}\), \(d_R(\hat{G},R_r)\) is zero if \(\hat{G}\in R_r\), otherwise the minimum Haversine distance from \(\hat{G}\) to \(R_r\).

\minisection{Existing metrics.}
\label{sec:standard_metrics}
Prior works report geolocalization accuracy as the percentage of predictions within \(1\), \(25\), \(200\), \(750\), and \(2500\) km of the ground-truth coordinate, corresponding approximately to street-, city-, region-, country-, and continent-level localization. Higher values indicate better performance. 
These metrics quantify absolute localization error, but do not capture whether predictions are stable under changes to the inserted person. We therefore complement them with race-based perturbation metrics that measure the magnitude and direction of counterfactual prediction shifts.

\subsection{Person-Origin Perturbation Indicators}
\label{sec:popi}
We introduce \emph{Person-Origin Perturbation Indicators} (POPI), a family of metrics that quantify whether inserting a person from a given race group systematically redirects a model’s prediction toward the geographic region associated with that group. Unlike standard geolocalization accuracy, POPI metrics are not intended to measure whether the model predicts the correct location. Instead, they measure race sensitivity: the degree to which the prediction changes in a direction consistent with the inserted person rather than with the fixed geographic scene.
Let \(s_i\) denote the index of the ground-truth source region of image \(x_i\), and let \(R_r\) denote the geographic support region associated with race group \(r\).
For each race-edited variant \(x_i^r\), we denote the corresponding prediction as \(\hat{G}_i^r = \mathcal{M}(x_i^r)\) and we define three complementary POPI metrics.

\minisection{POPI-Attack (POPI-A).} 
POPI-A measures how often inserting a person from race group \(r\) redirects the prediction toward the region associated with \(r\), starting from a background image whose original prediction does not already fall in that region. It is computed over samples satisfying \(\hat{G}_i \notin R_r\), as defined in Eq.~\eqref{eq:popi-a}:
\begin{equation}
        \mathrm{POPI\text{-}A}(r)
    =
    \frac{
    \sum_i \mathbbm{1}[\hat{G}_i \notin R_r]
          \mathbbm{1}[\hat{G}_i^{r} \in R_r]
    }{
    \sum_i \mathbbm{1}[\hat{G}_i \notin R_r]
    }.
    \label{eq:popi-a}
\end{equation}

A high POPI-A value indicates that the inserted person can steer the model toward the race-associated region, suggesting that the model is sensitive to foreground demographic cues that should be irrelevant to the ground-truth coordinate.

\minisection{POPI-Help (POPI-H).} 
POPI-H measures cases where the insertion of a person moves an initially incorrect prediction into the correct source region. This metric is computed only over background images whose original prediction fails at the source-region level, i.e., \(\hat{G}_i \notin R_{s_i}\), as defined in Eq.~\eqref{eq:popi-h}:
\begin{equation}
    \mathrm{POPI\text{-}H}(r)
    =
    \frac{
    \sum_i \mathbbm{1}[\hat{G}_i \notin R_{s_i}]
          \mathbbm{1}[\hat{G}_i^{r} \in R_{s_i}]
    }{
    \sum_i \mathbbm{1}[\hat{G}_i \notin R_{s_i}]
    }.
    \label{eq:popi-h}
\end{equation}
This metric captures cases in which the inserted person accidentally improves the prediction. Although such changes may increase apparent source-region accuracy, they still indicate dependence on a foreground intervention that is not part of the original geographic scene.

\minisection{POPI-Distraction (POPI-D).} 
POPI-D measures the opposite behavior: cases where a model initially predicts the correct source region for the background image, but the inserted person causes the prediction to leave that region. It is computed over samples satisfying \(\hat{G}_i \in R_{s_i}\), as defined in Eq.~\eqref{eq:popi-d}:
\begin{equation}
    \mathrm{POPI\text{-}D}(r)
    =
    \frac{
    \sum_i \mathbbm{1}[\hat{G}_i \in R_{s_i}]
          \mathbbm{1}[\hat{G}_i^{r} \notin R_{s_i}]
    }{
    \sum_i \mathbbm{1}[\hat{G}_i \in R_{s_i}]
    }.
    \label{eq:popi-d}
\end{equation}
High POPI-D indicates that the race intervention distracts the model from otherwise correct geographic evidence.

Overall, high values of POPI-A and POPI-D are undesirable: lower values indicate that a model is less susceptible to person-induced prediction shifts. POPI-H requires more careful interpretation. While a higher POPI-H may correspond to an apparent improvement in source-region accuracy, it still reflects dependence on the inserted person. Therefore, we report POPI-H as a diagnostic measure rather than as a fairness improvement.
To ensure that POPI is not confounded by varying region sizes, we also evaluate inverse-area-weighted aggregates; for GeoCLIP, this yields 33.7\% POPI-A, 22.4\% POPI-H, and 27.8\% POPI-D, confirming that demographic sensitivity is not an artifact of large geographic regions.

\subsection{Evaluated Models}
\label{sec:eval_models}
We evaluate seven geolocalization systems spanning two model families: dedicated image geolocalization models and general-purpose multimodal large language models.

\minisection{Dedicated geolocalization models.} We evaluate three recent image geolocalization systems. GeoCLIP~\cite{vivanco2023geoclip} learns a shared representation between image embeddings and GPS embeddings, enabling coordinate-based retrieval through contrastive alignment. Img2Loc~\cite{img2loc} follows a retrieval-augmented generation paradigm: it retrieves visually similar candidates and provides their coordinates to a multimodal language model, which then predicts a final location. G3~\cite{jia2024g3} similarly adopts a multi-stage geolocalization framework based on retrieval-augmented reasoning. These models represent different design choices in global visual geolocation, including coordinate embedding, retrieval-based candidate construction, and multimodal reasoning over geographic evidence.

\minisection{Multimodal large language models.} We also evaluate four open-weight MLLMs in a zero-shot setting: Qwen3-VL-8B-Instruct~\cite{qwen3technicalreport}, DeepSeek-VL-7B~\cite{lu2024deepseek}, InternVL3.5-8B-Instruct~\cite{wang2025internvl3}, and GLM-4.1V-9B~\cite{hong2025glm}. These models are not trained exclusively for image geolocalization, but their visual-language pretraining may allow them to exploit geographic cues such as scripts, road signs, vegetation, architecture, vehicles, terrain, climate, and cultural markers.
All MLLMs are evaluated with the same deterministic zero-shot prompt, in which the model is explicitly asked to return a single JSON object containing numeric latitude and longitude values, without explanations or chain-of-thought text. We parse the returned JSON object and discard outputs that cannot be converted into valid coordinates. The exact prompt is reported in Appendix~\ref{app:prompts}.
We additionally evaluate InternVL3.5-38B~\cite{wang2025internvl3}, Gemma3-27B-IT~\cite{gemma_2025}, and Qwen3VL-30B-A3B~\cite{qwen3technicalreport}; their results are reported in Appendix~\ref{app:large_mllms}.

\section{Experimental Results}
\label{sec:results}

\begin{table}
\caption{
\textbf{Overall results on \textsc{GeoBiaset}}. We report geolocalization accuracy at standard geodesic distance thresholds, corresponding approximately to street-, city-, region-, country-, and continent-level localization. The \emph{Edited} column indicates whether the image is an original background image (\xmark) or an edited image containing an inserted person (\cmark). Values in \textcolor{red}{red} denote the \textcolor{red}{relative accuracy drop} from background to edited images. POPI-A, POPI-H, and POPI-D measure race-induced prediction shifts; lower values indicate lower sensitivity to the inserted person.
}
\label{tab:main_table}
\centering
\resizebox{\columnwidth}{!}{%
\begin{tabular}{ccccccl|ccc}
\hline
Methods &
  Edited &
  \begin{tabular}[c]{@{}c@{}}Street\\ 1km\end{tabular} &
  \begin{tabular}[c]{@{}c@{}}City\\ 25km\end{tabular} &
  \begin{tabular}[c]{@{}c@{}}Region\\ 200km\end{tabular} &
  \begin{tabular}[c]{@{}c@{}}Country\\ 750km\end{tabular} &
  \begin{tabular}[l]{@{}c@{}}Continent\\ 2500km\end{tabular} &
  POPI-A ($\downarrow$) &
  POPI-H ($\downarrow$) &
  POPI-D ($\downarrow$)\\ \hline
  
\multirow{2}{*}{GeoCLIP} &
  \xmark &
  11.1 &
  35.6 &
  51.6 &
  68.8 &
  81.9 &
  \multirow{2}{*}{29.3} &
  \multirow{2}{*}{20.9} &
  \multirow{2}{*}{30.7} \\
 &
  \cmark &
  4.3\drop{61} &
  16.3\drop{54} &
  25.4\drop{51} &
  38.1\drop{45} &
  55.3\otherdrop{33} &
   &
   &
   \\

\hline
\multirow{2}{*}{Img2Loc} &
  \xmark &
  8.2 &
  28.0 &
  37.0 &
  53.1 &
  70.3 &
  \multirow{2}{*}{34.5} &
  \multirow{2}{*}{24.3} &
  \multirow{2}{*}{36.2} \\
 &
  \cmark &
  2.2\drop{73} &
  7.5\drop{73} &
  11.8\drop{68} &
  21.8\drop{59} &
  37.5\otherdrop{47} &
   &
   &
   \\

\hline
\multirow{2}{*}{G3} &
  \xmark &
  16.6 &
  45.8 &
  55.7 &
  70.3 &
  81.6 &
  \multirow{2}{*}{28.1} &
  \multirow{2}{*}{20.0} &
  \multirow{2}{*}{29.4} \\
 &
  \cmark &
  3.6\drop{78} &
  13.3\drop{71} &
  19.0\drop{66} &
  30.7\drop{56} &
  47.8\otherdrop{41} &
   &
   &
   \\

\hline
\multirow{2}{*}{DeepSeek-VL} &
  \xmark &
  0.9 &
  5.5 &
  15.5 &
  41.1 &
  64.7 &
  \multirow{2}{*}{18.2} &
  \multirow{2}{*}{19.3} &
  \multirow{2}{*}{18.0} \\
 &
  \cmark &
  {0.7}\drop{18} &
  {5.2}\drop{7} &
  {12.9}\drop{17} &
  {31.3}\drop{24} &
  {51.5}\otherdrop{20} &
   &
   &
   \\
\hline
\multirow{2}{*}{InternVL3.5} &
  \xmark &
  0.9 &
  5.5 &
  12.8 &
  34.7 &
  58.0 &
  \multirow{2}{*}{24.1} &
  \multirow{2}{*}{24.6} &
  \multirow{2}{*}{24.1} \\
 &
  \cmark &
  {0.8}\drop{3} &
  {4.6}\drop{18} &
  {9.7}\drop{24} &
  {23.4}\drop{33} &
  {43.2}\otherdrop{25} &
   &
   &
   \\

\hline
\multirow{2}{*}{Qwen3-VL} &
  \xmark &
  3.2 &
  24.2 &
  38.5 &
  62.7 &
  79.3 &
  \multirow{2}{*}{17.8} &
  \multirow{2}{*}{17.6} &
  \multirow{2}{*}{17.8} \\
 &
  \cmark &
  {2.6}\drop{19} &
  {19.2}\drop{21} &
  {30.0}\drop{22} &
  {49.6}\drop{21} &
  {65.3}\otherdrop{18} &
   &
   &
   \\

\hline
\multirow{2}{*}{GLM-4.1V} &
  \xmark &
  3.5 &
  26.5 &
  47.2 &
  71.4 &
  82.8 &
  \multirow{2}{*}{16.2} &
  \multirow{2}{*}{14.8} &
  \multirow{2}{*}{16.4} \\
 &
  \cmark &
  {2.5}\drop{29} &
  {21.3}\drop{20} &
  {37.4}\drop{21} &
  {56.9}\drop{20} &
  {69.2}\otherdrop{16} &
   &
   &
   \\
      \hline
  
\end{tabular}%
}
\end{table}

\subsection{Overall Geolocalization Performance and Race-Induced Perturbations}\label{sec:results_overall}
Tab.~\ref{tab:main_table} reports results for all 7 models. From these we draw three conclusions.

First, inserting a person consistently degrades accuracy across all models and all geographic scales, most sharply for task-specific systems: GeoCLIP drops from 81.9\% to 55.3\% at continent level, Img2Loc from 70.3\% to 37.5\%, and G3 from 81.6\% to 47.8\%. Even models built for geolocalization are sensitive to foreground human subjects despite the background and GPS coordinate being unchanged.

Second, dedicated models are more strongly affected than general-purpose MLLMs. At continent level, the relative drop is 33--47\% for the former and 16--25\% for the latter. Higher background-image accuracy does not imply greater robustness to demographic interventions: MLLMs have lower fine-grained accuracy but their predictions are less sharply displaced by the inserted person.

Third, POPI shows the degradation is often a directional shift. POPI-A exceeds 28\% for all dedicated models and ranges from 16.2\% to 24.1\% for MLLMs. Thus, in a substantial fraction of cases, inserting a person from a given race causes the model to predict a coordinate inside the region associated with that race. 
Overall, these results support the hypothesis that current geolocalization systems can exploit race-associated foreground cues as spurious geographic shortcuts.

\subsection{Causal Isolation of Foreground Demographic Cues}
\label{sec:causal_isolation}
The comparison between original and edited images may conflate sensitivity to the inserted subject with two additional effects: the subject physically occludes geographic evidence, and the image-editing model may alter portions of the background. We therefore perform a controlled masking experiment using SAM3~\cite{carion2025sam3segmentconcepts}. For each edited image, we segment the inserted person and replace the corresponding pixels with three masking conditions: black, uniform gray, and Gaussian blur. Crucially, the same variant-specific mask is also applied to the matched original image, so that the amount, position, and shape of the occlusion are identical in each pair.

Once the subject is removed, edited and original backgrounds achieve very similar geolocalization performance. At the continent level, black masking yields $73.4\%$ accuracy on the masked originals and $72.7\%$ on the corresponding masked edits, while gray masking yields $73.6\%$ and $71.7\%$, respectively. In contrast, accuracy on the unmasked edited images is only $55.3\%$, compared with $81.9\%$ on the original backgrounds. Thus, foreground occlusion is a measurable source of degradation, but neither occlusion nor generation-induced background drift explains most of the observed accuracy loss.

Combining this control with the face-obfuscation experiment of Sec.~\ref{sec:results_anonymized}, the $26.6$ point continent-level drop can be approximately decomposed into $8.5$ points due to scene occlusion, $0.7$ points due to generation-induced background changes, $9.0$ points due to non-facial subject cues, and $8.4$ points due to facial appearance. These results support the conclusion that the visible inserted subject, rather than background corruption alone, is the dominant source of the observed degradation. Full results and implementation details are reported in Appendix~\ref{app:causal_isolation}.

\subsection{Landmark Sensitivity and Attribution Analysis}
\label{sec:results_landmark_attention}

\begin{figure}[t]
    \centering

    \begin{minipage}[c]{0.42\textwidth}
        \centering
        \begin{tikzpicture}
        \begin{axis}[
            ybar,
            width=\linewidth,
            height=0.75\linewidth,
            bar width=15pt,
            enlarge x limits=0.48,
            ylabel={\%},
            ylabel style={font=\small},
            symbolic x coords={POPI-H, POPI-D},
            xtick=data,
            xticklabels={POPI-H, POPI-D},
            tick label style={font=\small},
            x tick label style={font=\small},
            yticklabel style={font=\small},
            ymin=0,
            ymax=42,
            ytick={0,10,20,30,40},
            ymajorgrids=true,
            grid style={dashed, gray!30},
            axis line style={black!70},
            nodes near coords,
            nodes near coords style={
                font=\scriptsize,
                /pgf/number format/fixed,
                /pgf/number format/precision=1
            },
            every node near coord/.append style={yshift=1.5pt},
            legend style={
                at={(0.5,1.05)},
                anchor=south,
                legend columns=2,
                font=\scriptsize,
                draw=none,
                fill=none,
                column sep=4pt
            },
            legend image code/.code={
                \draw[#1, draw=black] (0cm,-0.08cm) rectangle (0.16cm,0.08cm);
            }
        ]
        \addplot[fill=ibmBlue, draw=black]
            coordinates {(POPI-H, 9.17) (POPI-D, 25.97)};
        \addplot[fill=ibmOrange, draw=black]
            coordinates {(POPI-H, 27.13) (POPI-D, 33.20)};
        \legend{Landmark, Non-landmark}
        \end{axis}
        \end{tikzpicture}
        \label{fig:popi_delta_accuracy}
    \end{minipage}%
    \hfill
    \begin{minipage}[c]{0.56\textwidth}
        \centering
        \begin{subfigure}[t]{0.315\linewidth}
            \centering
            \includegraphics[width=\linewidth]{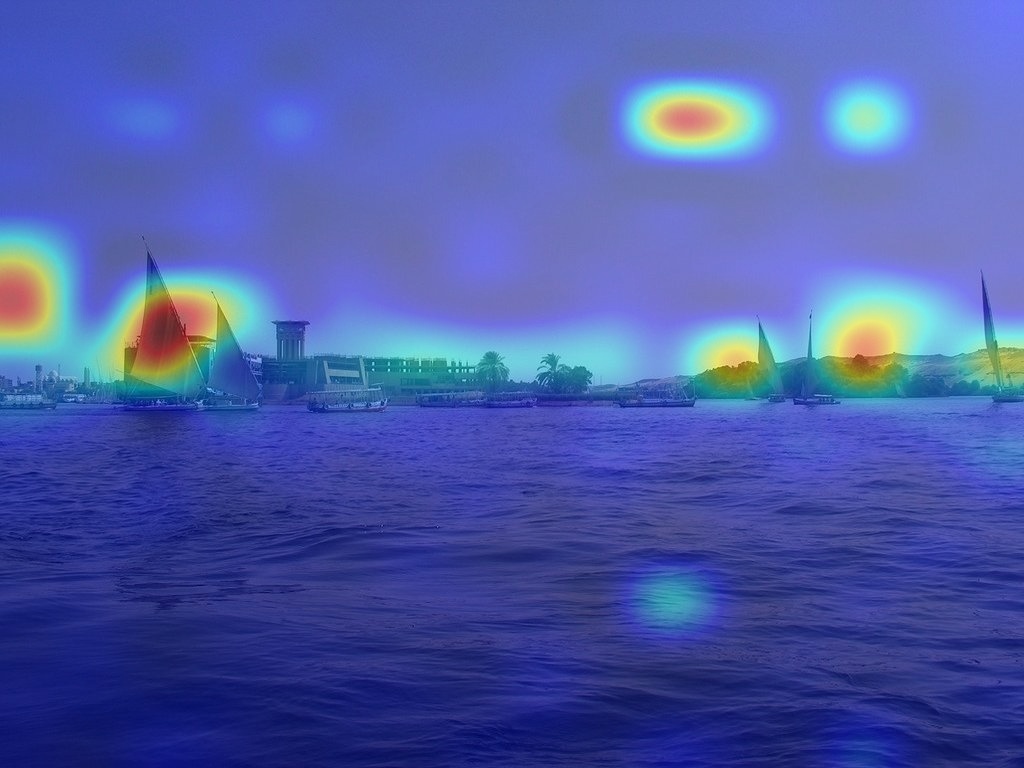}
            \caption{Landmark}
        \end{subfigure}\hfill
        \begin{subfigure}[t]{0.315\linewidth}
            \centering
            \includegraphics[width=\linewidth]{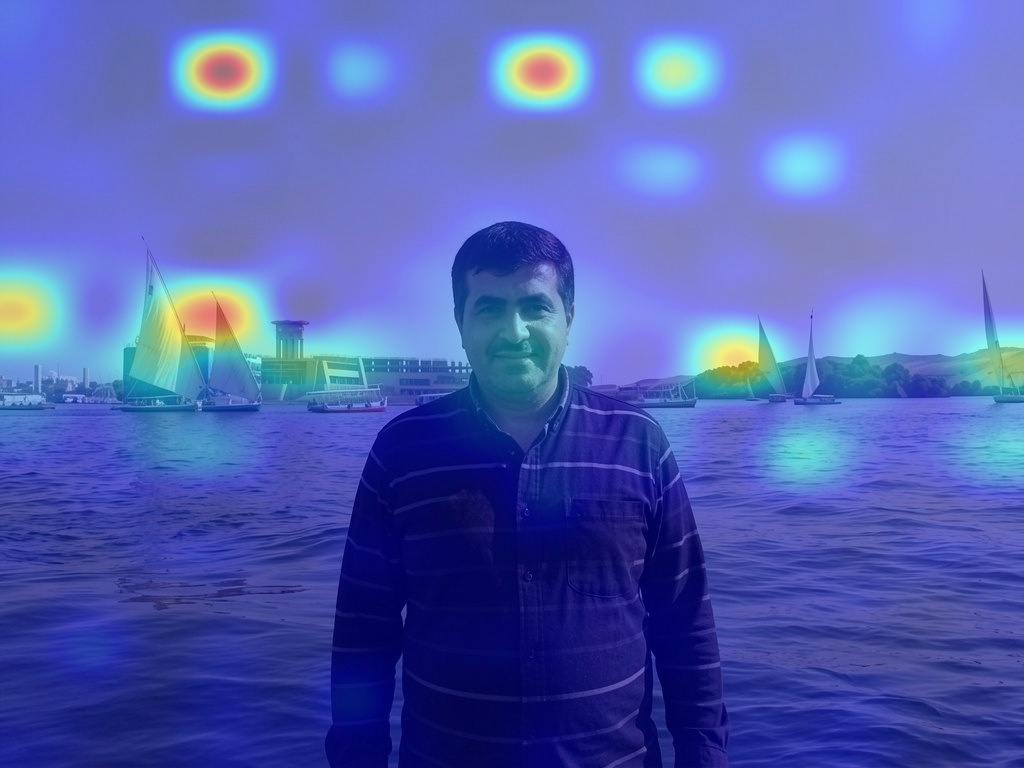}
            \caption{Help}
        \end{subfigure}\hfill
        \begin{subfigure}[t]{0.315\linewidth}
            \centering
            \includegraphics[width=\linewidth]{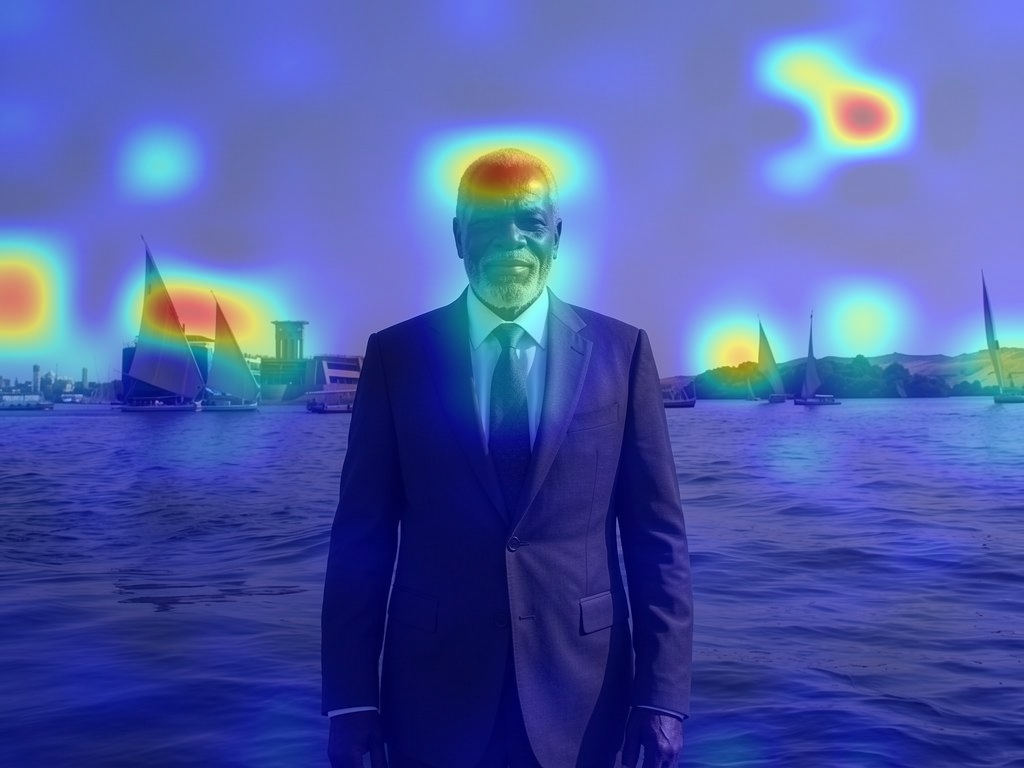}
            \caption{Distraction}
        \end{subfigure}

        \vspace{3pt}

        \begin{subfigure}[t]{0.315\linewidth}
            \centering
            \includegraphics[width=\linewidth]{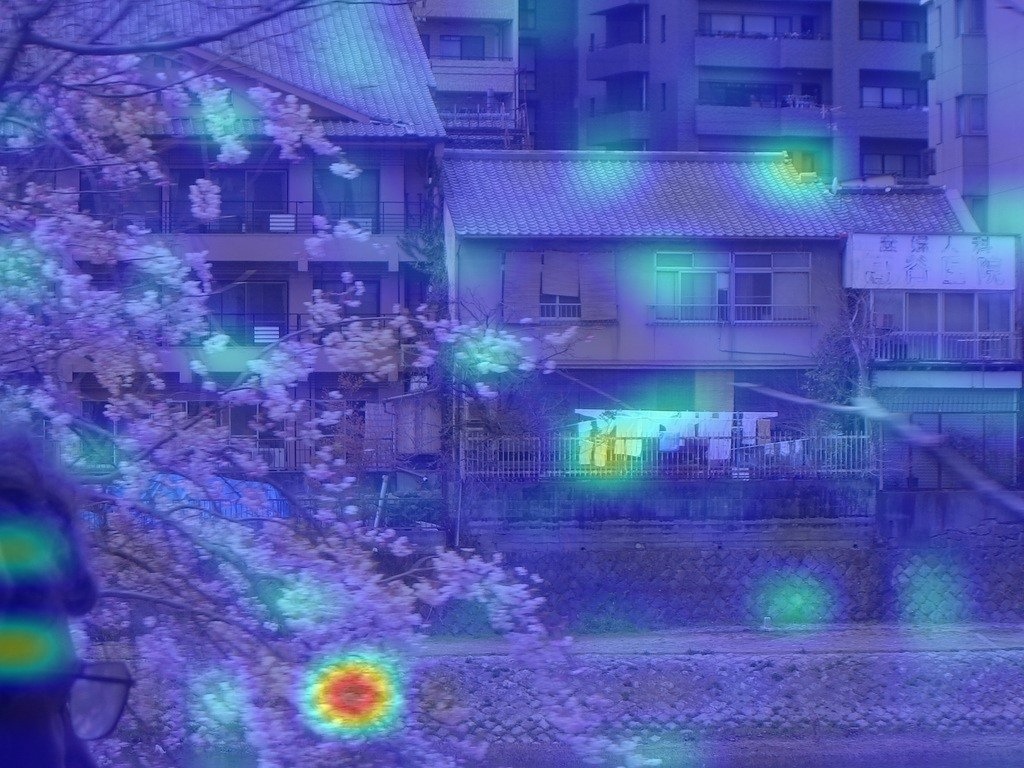}
            \caption{Non-landmark}
        \end{subfigure}\hfill
        \begin{subfigure}[t]{0.315\linewidth}
            \centering
            \includegraphics[width=\linewidth]{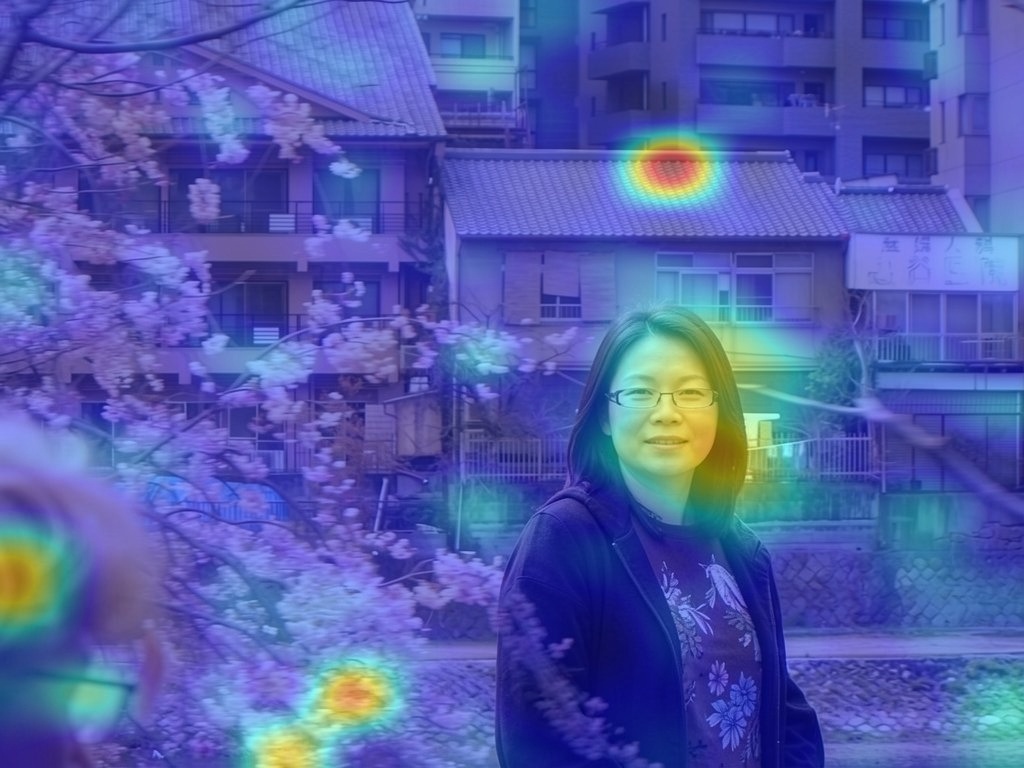}
            \caption{Help}
        \end{subfigure}\hfill
        \begin{subfigure}[t]{0.315\linewidth}
            \centering
            \includegraphics[width=\linewidth]{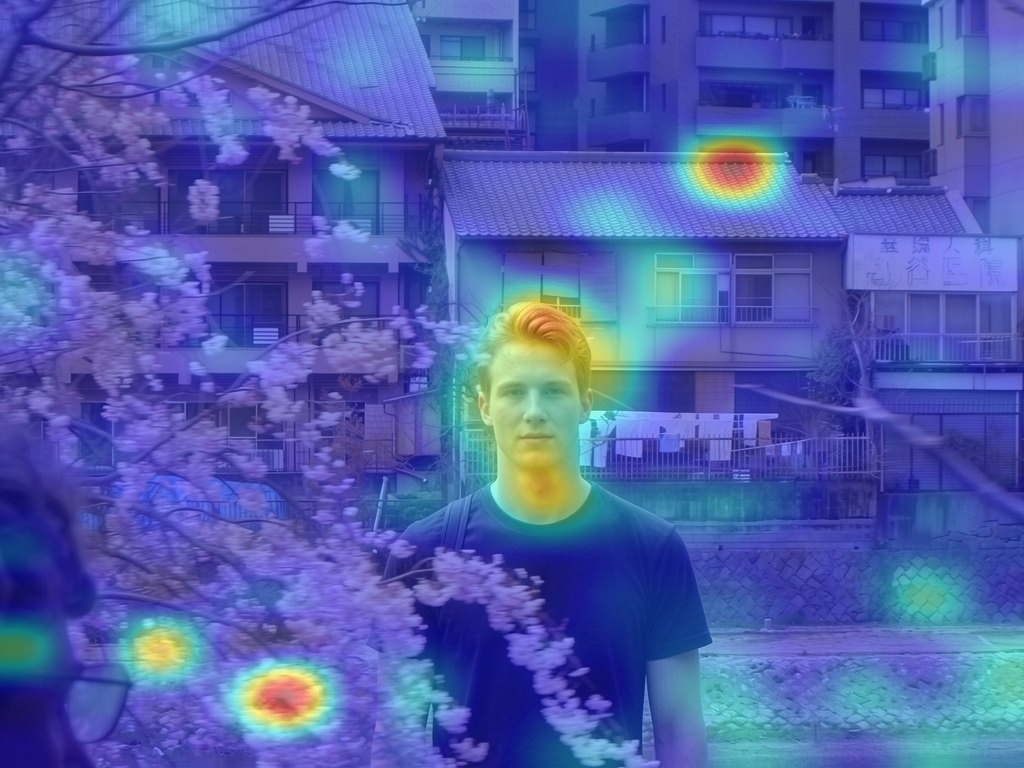}
            \caption{Distraction}
        \end{subfigure}
        \label{fig:attention_map_landmark_non_landmark}
    \end{minipage}

    \caption{Landmark sensitivity and attribution analysis for GeoCLIP.
    \textbf{Left}: POPI-H and POPI-D for landmark and non-landmark images.
    \textbf{Right}: representative gradient-weighted attention rollout maps showing that non-landmark scenes assign more relevance to the inserted person, consistent with larger prediction shifts.}
    \label{fig:popi_delta_and_attention_maps}
\end{figure}
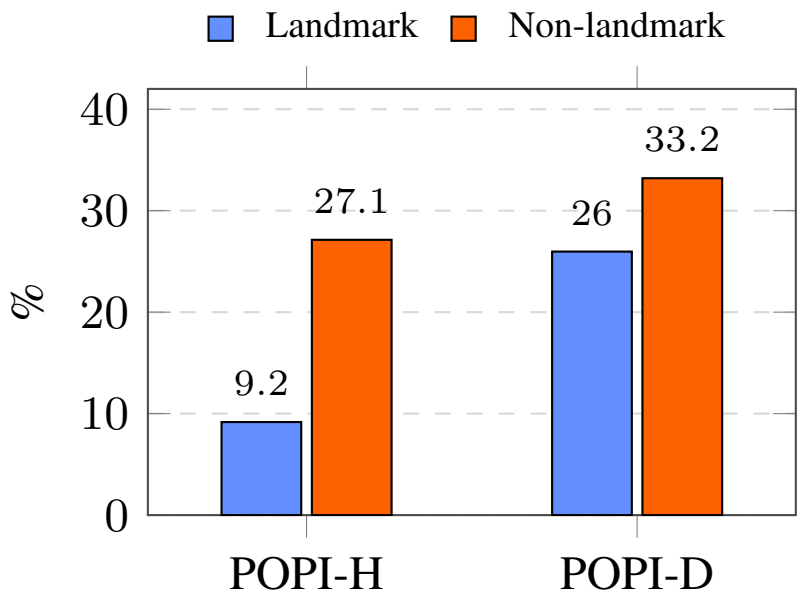
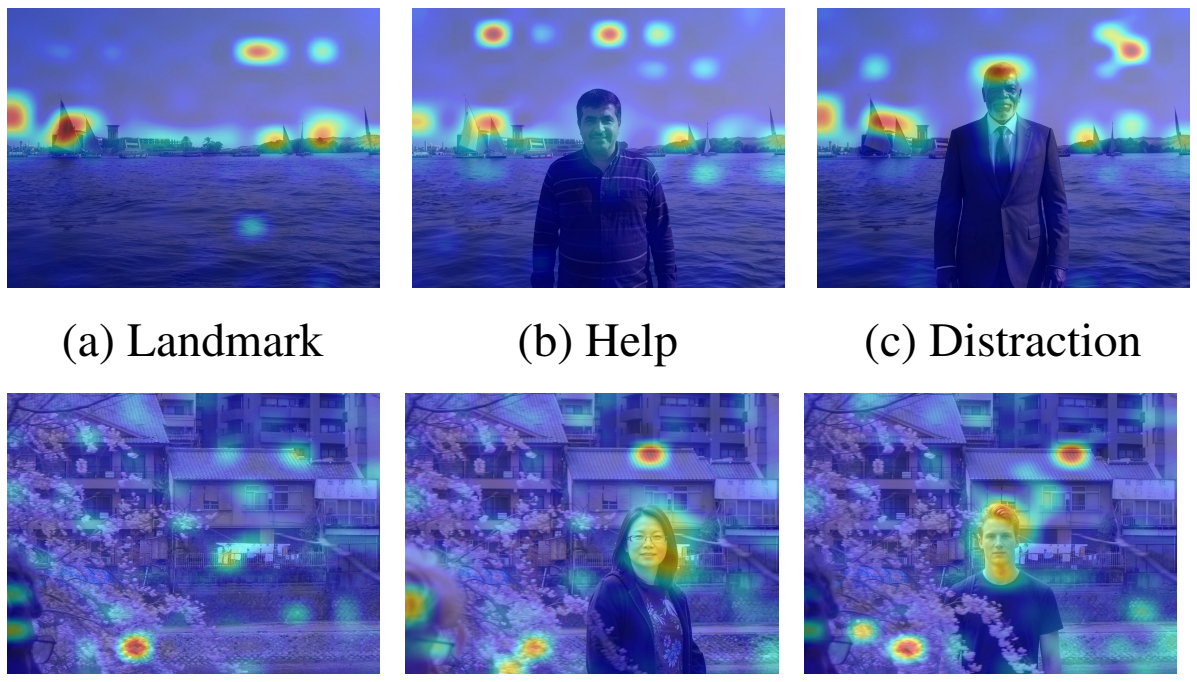
We further analyze whether the presence of a landmark moderates the effect of the inserted person. Landmarks provide highly localized visual evidence and may therefore make geolocalization models less dependent on foreground demographic cues. Conversely, in non-landmark scenes, where the geographic evidence is weaker or more ambiguous, models may rely more heavily on spurious cues introduced by the edited subject.
Fig.~\ref{fig:popi_delta_and_attention_maps} reports this analysis for GeoCLIP by combining aggregate POPI sensitivity with representative attribution maps. The left panel compares \(\Delta_{\mathrm{Acc}}\) under POPI-H and POPI-D for landmark and non-landmark images. The results indicate that landmarks attenuate, but do not eliminate, race-induced perturbations. 
For POPI-H, the aggregate accuracy change is substantially larger in non-landmark images than in landmark images, increasing from 9.2\% to 27.1\%. A similar pattern holds for POPI-D, which increases from 26.0\% in landmark scenes to 33.2\% in non-landmark scenes. This suggests that when distinctive geographic evidence is weak or absent, the model becomes more susceptible to foreground demographic cues.

To better understand this behavior, we visualize GeoCLIP evidence using gradient-weighted attention rollout over CLIP vision transformer~\cite{abnar-zuidema-2020-quantifying, chefer2021transformer}. 
The attention maps in Fig.~\ref{fig:popi_delta_and_attention_maps} provide qualitative evidence consistent with the aggregate trend. In landmark scenes, the model continues to assign relevance to characteristic location-specific structures, although the inserted person also receives attention. In non-landmark scenes, relevance is more concentrated on the inserted subject, and the prediction is more likely to shift after the demographic intervention. Overall, these results suggest that strong geographic cues can partially stabilize model predictions, but they do not fully prevent race-associated foreground information from influencing geolocalization decisions.

\subsection{Effect of Race-Associated Attire}
\label{sec:results_attire}

\begin{wraptable}{r}{0.5\columnwidth}
\vspace{-1.2cm}
\centering
\caption{
\textbf{Effect of race-associated attire on GeoCLIP accuracy.} Values report the accuracy change between attire-conditioned and standard edited images. Negative values indicate accuracy drops; darker cells denote larger drops.
}
\resizebox{0.5\columnwidth}{!}{%
\begin{tabular}{lrrrrr}
\hline
Race & 1km & 25km & 200km & 750km & 2500km \\ \hline
Black           & \hB{-2.2} & \hC{-4.7} & \hE{-9.9} & \hF{-12.4} & \hF{-15.5} \\
East Asian      & \hA{-1.2} & \hC{-4.1} & \hD{-7.0} & \hE{-8.7}  & \hE{-10.6} \\
Indian          & \hA{-0.6} & \hB{-1.9} & \hC{-5.0} & \hC{-3.6}  & \hC{-3.6}  \\
Latino Hispanic & \hA{0.0}  & \hC{-3.6} & \hC{-3.9} & \hD{-5.1}  & \hD{-5.4}  \\
Middle Eastern  & \hB{-1.6} & \hB{-2.0} & \hC{-3.5} & \hB{-1.6}  & \hA{+0.4}   \\
Southeast Asian & \hA{-0.1} & \hC{-3.1} & \hD{-6.6} & \hD{-7.3}  & \hE{-9.0}  \\
White           & \hB{-2.6} & \hC{-3.8} & \hD{-7.3} & \hC{-4.2}  & \hB{-1.6}  \\ \hline
\end{tabular}
}
\vspace{-0.4cm}
\label{tab:acc-delta-heatmap}
\end{wraptable}
We next test whether race-associated clothing cues further affect predictions. The inserted person is generated with clothing sampled from race- and gender-specific lists (Appendix~\ref{app:garment_list}). This intervention tests reliance not only on facial appearance or body-level attributes, but also on culturally associated visual cues spuriously correlated with geography.

Tab.~\ref{tab:acc-delta-heatmap} reports the accuracy change for GeoCLIP across distance thresholds. Race-associated attire substantially amplifies errors at coarse scales: a Black subject in race-associated attire reduces continent-level accuracy by 15.5\%; East Asian and Southeast Asian reduce it by 10.6\% and 9.0\% respectively. Clothing acts as a \textit{strong geographic prior}, 
but the effect is not uniform: some conditions produce smaller drops, and the Middle Eastern condition does not decrease continent-level accuracy in this setting. This heterogeneity suggests attire-based perturbations interact with both the background content and the model's learned associations between appearance, culture, and geography. Overall, the results suggest that race-associated clothing is therefore an important factor in demographic geolocation bias and should be evaluated separately from facial appearance. Results for the remaining models are in Appendix~\ref{suppl:results_attire}.

\subsection{Ablation on Face-Obfuscated Images}
\label{sec:results_anonymized}
We test whether the perturbation is driven by facial appearance by applying face obfuscation to \textsc{GeoBiaset}, which preserves the rest of the image while blurring the inserted person's face. If the drop were mainly caused by facial identity, obfuscation should largely remove it.

Tab.~\ref{tab:geoclip-anonymized-delta} reports the change in GeoCLIP accuracy with face obfuscation. Obfuscation slightly weakens the perturbation in several cases, but the overall drop remains substantial for most groups. 
\begin{wraptable}{r}{0.5\columnwidth}
\centering
\caption{
\textbf{Effect of face obfuscation on GeoCLIP accuracy.} Values report the accuracy gain after applying face blurring to the edited images, relative to the corresponding non-obfuscated edited images. Positive values indicate that obfuscation improves accuracy, suggesting reduced sensitivity to explicit facial cues. Darker cells denote larger gains.
}
\resizebox{0.5\columnwidth}{!}{%
\begin{tabular}{lrrrrr}
\hline
Race & 1km & 25km & 200km & 750km & 2500km \\ \hline
Black           & \oB{+1.3}  & \oC{+4.9} & \oD{+5.7} & \oE{+8.3}  & \oE{+8.8}  \\
East Asian      & \oB{+1.4}  & \oC{+4.0} & \oD{+6.1} & \oE{+8.6}  & \oE{+9.2}  \\
Indian          & \oB{+2.7}  & \oD{+5.9} & \oD{+5.4} & \oE{+7.5}  & \oE{+8.8}  \\
Latino Hispanic & \oB{+1.0}  & \oC{+4.3} & \oD{+6.2} & \oF{+9.4}  & \oF{+11.7} \\
Middle Eastern  & \oA{+0.3}  & \oD{+5.3} & \oD{+6.5} & \oF{+11.0} & \oF{+10.7} \\
Southeast Asian & \oB{+1.4}  & \oC{+4.1} & \oE{+7.2} & \oE{+9.0}  & \oE{+7.9}  \\
White           & \oA{0.0} & \oB{+1.3} & \oB{+2.2} & \oC{+3.2}  & \oB{+1.7}  \\ \hline
\end{tabular}
}
\label{tab:geoclip-anonymized-delta}
\end{wraptable}
Face obfuscation increases the localization accuracy of edited images especially for Latino Hispanic, Middle Eastern, East Asian, Indian, and Black, ranging between 8.8--11.7\%. The White group shows a smaller difference, so the effect is not uniform.

Results indicate that facial appearance alone does not explain the model's sensitivity. Even with the face blurred, the model remains affected by non-facial cues such as clothing, skin tone outside the face region, body appearance, pose, and contextual interactions between the inserted person and the background. Demographic bias in geolocalization can persist under face anonymization, which matters for both privacy-preserving evaluation and responsible deployment. Face obfuscation is at best a first-aid mitigation against severe degradation in biased models.

\section{Ethical Considerations and Limitations}
\label{sec:ethics_limitations}

This work studies demographic bias in a task with clear dual-use implications. Geolocalization can support journalism, historical analysis, environmental monitoring, geography education, and open-source verification, but it can also enable surveillance, stalking, military targeting, or identification of sensitive locations. Bias in such systems can amplify harm if errors or shifts are unevenly distributed across demographic groups or regions. \textsc{GeoBiaset} is intended as an auditing benchmark for measuring and mitigating these risks, not as a tool for improving harmful uses of geolocation technology.

Our design also simplifies human identity. We follow the FairFace race taxonomy and binary gender annotations to obtain a controlled, reproducible benchmark, but these categories do not capture the full complexity of race, ethnicity, gender identity, migration, diaspora, or cultural belonging. Race-associated regions and clothing prompts are operational constructs for measurement and should not be read as intrinsic, exhaustive, or normative mappings. A high POPI score indicates model sensitivity to race-associated geographic shortcuts, not demographic truth.

Finally, \textsc{GeoBiaset} relies on synthetic image editing rather than masked inpainting. While the source GPS coordinate is fixed and the geographic scene is largely preserved, the editing process does not guarantee pixel-level background identity. Full-reference measurements on the unmasked background reveal measurable generation-induced changes, including occasional lighting shifts and localized artifacts. The pipeline also does not
explicitly constrain subject scale or placement, and our automated quality control cannot guarantee geometric consistency or replace formal human evaluation. However, the controlled masking experiment in Sec.~\ref{sec:causal_isolation} shows that these effects do not primarily explain the observed geolocalization degradation. The benchmark is additionally limited to 329 source backgrounds and focuses on foreground people; future work should broaden geographic coverage, strengthen intervention control, and study other
sources of geolocalization bias.

\section{Conclusion}
\label{sec:conclusion}

We introduced \textsc{GeoBiaset}, a counterfactual benchmark for auditing demographic bias in worldwide image geolocalization, together with POPI metrics for measuring directional prediction shifts under demographic interventions. By decoupling foreground human attributes from the geographic scene, the benchmark exposes a failure mode that standard geodesic accuracy cannot capture.

Across dedicated geolocalization models and zero-shot multimodal large language models, we find that predictions are often unstable under changes to the inserted person and can shift toward race-associated geographic regions. These results show that high benchmark accuracy does not necessarily imply robust geographic reasoning. We hope \textsc{GeoBiaset} encourages future geolocalization systems to report counterfactual robustness alongside accuracy and to reduce reliance on demographic shortcuts.

\clearpage

{
\small
\bibliographystyle{plain}
\bibliography{biblio}
}

\clearpage
\appendix
\begin{appendices}
\section{Full Counterfactual \textsc{GeoBiaset} example}
\label{app:full_grids}
\begin{figure}[hb]
\centering
\includegraphics[width=0.8\textwidth]{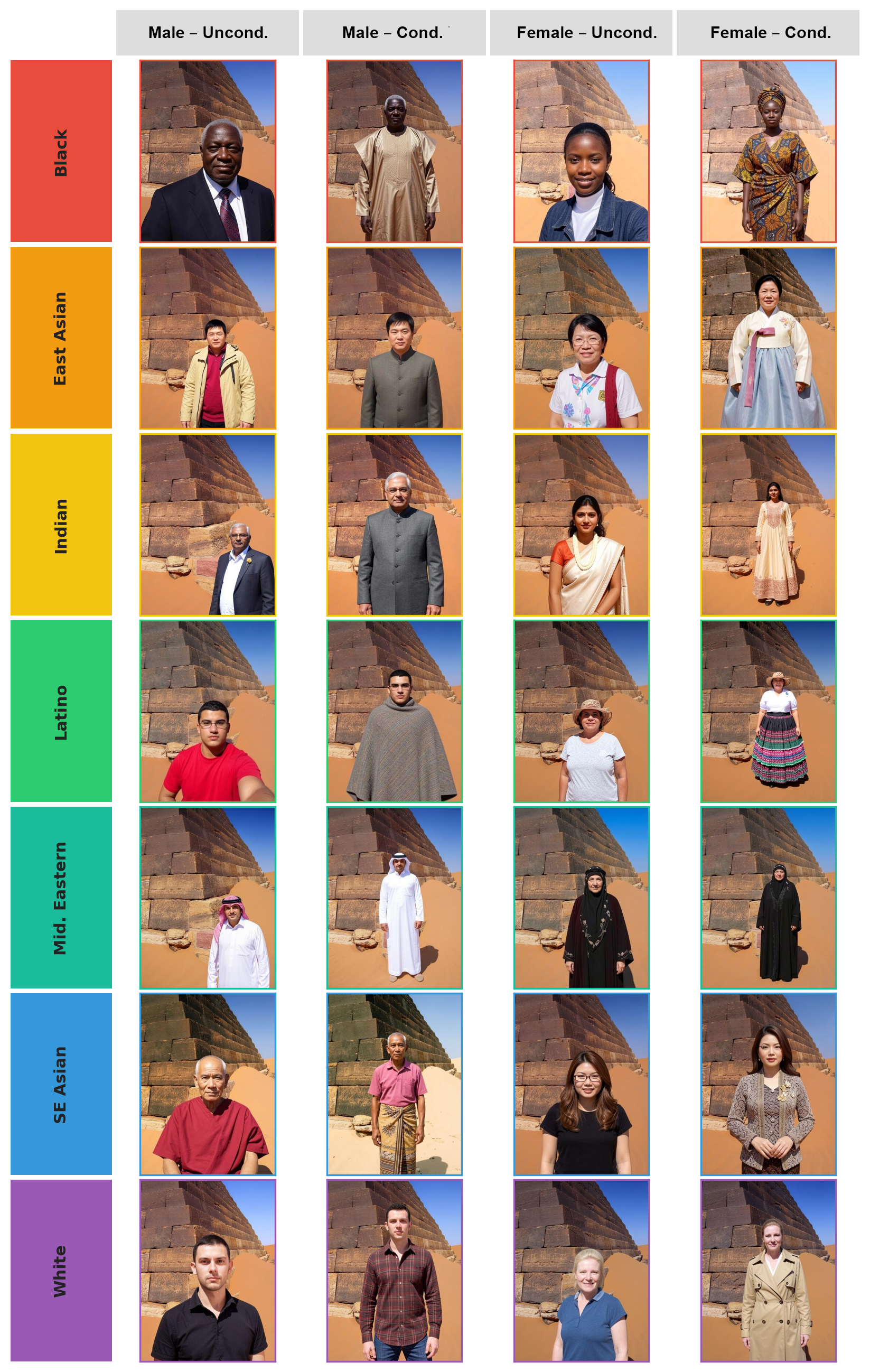}
\caption{Full \textsc{GeoBiaset} counterfactual grid for a \textbf{landmark} scene with all 28 edited variants across seven races (rows) and four gender-attire conditions (columns), sharing the same ground-truth GPS coordinate; subject composition and background appearance may vary across edits.}

\label{fig:full_grid_landmark}
\end{figure}

\begin{figure}
    \centering
    \includegraphics[width=0.95\textwidth]{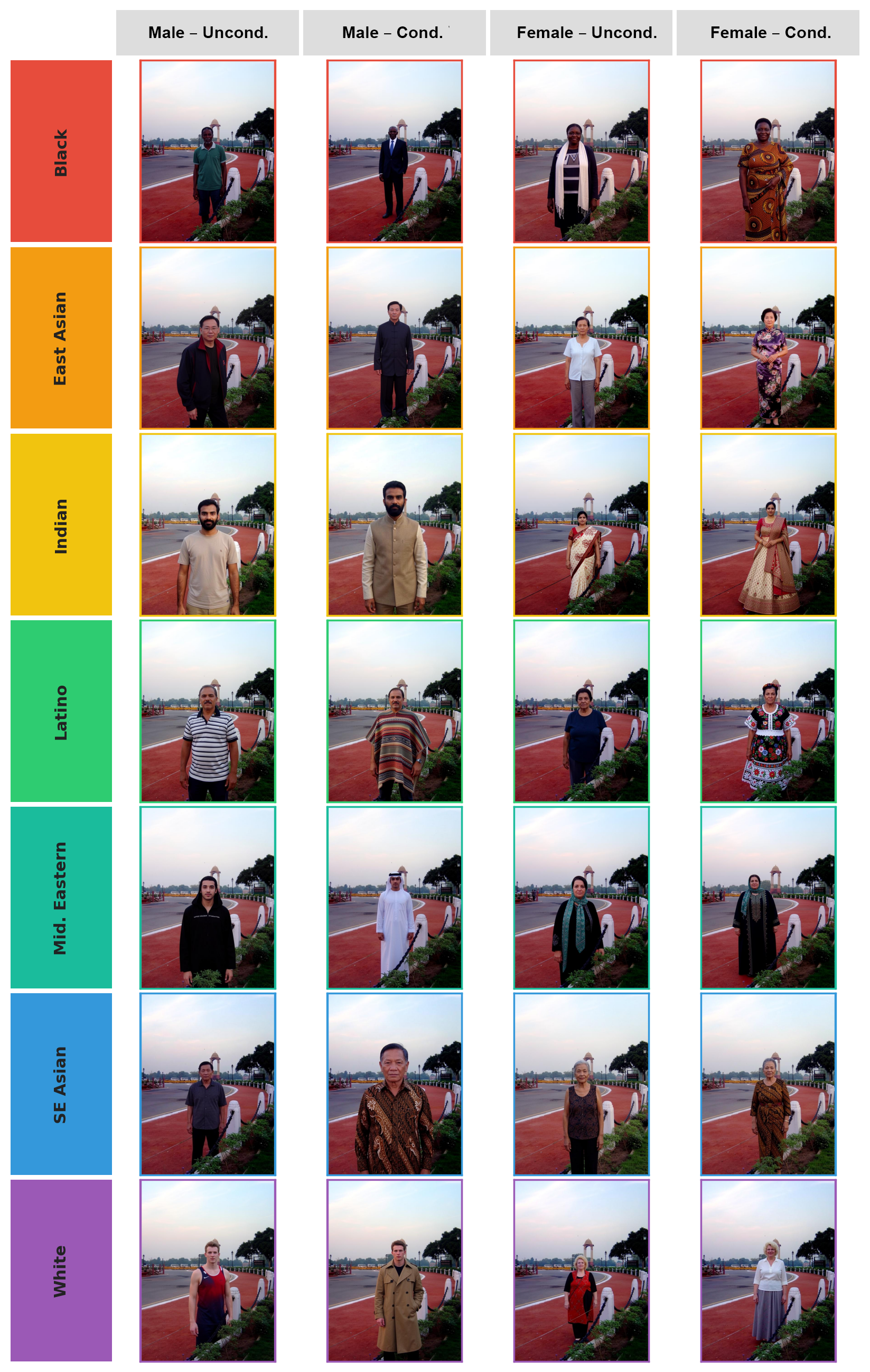}
    \caption{Full \textsc{GeoBiaset} counterfactual grid for a \textbf{non-landmark} scene with all 28 edited variants across seven races (rows) and four gender-attire conditions (columns), sharing the same ground-truth GPS coordinate; subject composition and background appearance may vary across edits.}

    \label{fig:full_grid_nonlandmark}
\end{figure}

To complement the quantitative analysis in Sec.~\ref{sec:results}, we visualize the complete set of \textsc{GeoBiaset} counterfactual variants generated from a single background image. Fig.~\ref{fig:full_grid_landmark} shows a landmark scene and Fig.~\ref{fig:full_grid_nonlandmark} a non-landmark scene. 


\section{Prompts}\label{app:prompts}
\subsection{Image editing prompt}\label{prompt:image_editing}
To generate the dataset images we use the following prompt:

\begin{tcolorbox}[
    enhanced,
    colback=gray!5,
    colframe=gray!50,
    boxrule=0.4pt,
    arc=2pt,
    left=8pt, right=8pt, top=6pt, bottom=6pt,
    fontupper=\itshape\small,
    title=\textbf{Image Editing Prompt},
    coltitle=black,
    colbacktitle=gray!20,
    attach boxed title to top left={xshift=6pt, yshift=-2pt},
    boxed title style={colframe=gray!50, sharp corners, boxrule=0.3pt},
    breakable
]
Generate a [\keyword{RACE}] [\keyword{GENDER}] person of age [\keyword{AGE}] years in image 1 looking like the person from image 2. Make the generated person look as similar as possible to the person in image 2. Make the person pretty. Keep the lighting and the background as in image 1. (Make the generated person face the camera: 0.5). Give the picture a natural feeling, as if it was taken by a camera. (Make the generated person look like they belong in the scene: 1.4). Keep a natural pose. (Do not make more than two arms, legs or faces).
\end{tcolorbox}

\subsection{Clothing prompt}\label{prompt:clothing}

\begin{tcolorbox}[
    enhanced,
    colback=gray!5,
    colframe=gray!50,
    boxrule=0.4pt,
    arc=2pt,
    left=8pt, right=8pt, top=6pt, bottom=6pt,
    fontupper=\itshape\small,
    title=\textbf{Clothing Prompt},
    coltitle=black,
    colbacktitle=gray!20,
    attach boxed title to top left={xshift=6pt, yshift=-2pt},
    boxed title style={colframe=gray!50, sharp corners, boxrule=0.3pt},
    breakable
]
The person is wearing typical [\keyword{GENDER ADJ}] [\keyword{CLOTHING SAMPLE}]: 1.4.
\end{tcolorbox}

\label{app:garment_list}
Here, \keyword{GENDER ADJ} is set to ``masculine'' for male subjects and ``feminine'' for female subjects to reduce ambiguity for garments that may be worn across genders. \keyword{CLOTHING SAMPLE} is drawn from a curated list of typical garments generated with Claude Sonnet 4.6, reported in Tab. \ref{tab:garment-list}

\begin{table}[h]
\centering
\small
\renewcommand{\arraystretch}{1.3}
\caption{Curated garment list per race, split by gender adjective.}
\begin{tabular}{p{0.13\textwidth} p{0.38\textwidth} p{0.38\textwidth}}
\toprule
\textbf{Race} & \textbf{Masculine garments} & \textbf{Feminine garments} \\
\midrule
White & suit and tie, jeans and t-shirt, chinos with polo shirt, bomber jacket, cable-knit sweater, trench coat, flannel shirt with jeans, tuxedo & sundress, jeans and blouse, pencil skirt with top, trench coat, wrap dress, cardigan with slacks, maxi skirt with blouse, cocktail dress \\
Black & dashiki, suit and tie, streetwear hoodie and joggers, agbada, senator suit, boubou, kente cloth shirt, kaftan & ankara wrap dress, dashiki blouse with trousers, kente cloth dress, gele headwrap with boubou, iro and buba, lappa skirt with top, kitenge dress, adire wrap dress \\
Southeast Asian & batik shirt, sarong with shirt, barong tagalog, longyi, baju melayu, pha nung, puso ng saging, teluk belanga & kebaya, ao dai, sarong with blouse, baro't saya, baju kurung, pha sin wrap skirt, inabel dress, nyonya kebaya \\
East Asian & changshan, hanbok (jeogori and baji), hakama with kimono, modern suit, mandarin collar jacket, haori, magua jacket, durumagi overcoat & qipao, hanbok (jeogori and chima), kimono, hanfu, furisode, jeogori with maxi chima, ruqun, uchikake \\
Latino/Hispanic & guayabera shirt, charro suit, poncho with trousers, linen shirt and slacks, ruana cloak, jorongo poncho, montuno shirt, bombacha trousers & huipil, pollera, china poblana, floral wrap skirt with blouse, mestiza dress, cholita layered skirt, jorongo shawl with dress, rebozo wrap with blouse \\
Middle Eastern & thobe (dishdasha), keffiyeh with agal, shalwar kameez, jalabiya, bisht cloak, sirwal trousers with tunic, jubba, egal with ghutra & abaya, kaftan, jilbab with hijab, thobe al-nashal, farasha dress, jalabiya with tarha, bisht evening cloak, shayla with maxi dress \\
Indian & kurta with churidar, sherwani, dhoti with angavastram, nehru jacket, achkan coat, bandhgala suit, lungi with shirt, pathani suit & saree, salwar kameez, lehenga choli, anarkali suit, churidar with kurti, gharara, sharara with dupatta, ikat silk dress \\
\bottomrule
\end{tabular}
\label{tab:garment-list}
\end{table}

\clearpage
\subsection{Quality control prompt}\label{prompt:quality_control}

For the quality-control model \(\Psi_g\), we use Qwen3-VL-8B-Instruct~\cite{qwen3technicalreport}. The model is prompted with:

\begin{tcolorbox}[
    enhanced,
    colback=gray!5,
    colframe=gray!50,
    boxrule=0.4pt,
    arc=2pt,
    left=8pt, right=8pt, top=6pt, bottom=6pt,
    fontupper=\itshape\small,
    title=\textbf{Quality Control Prompt},
    coltitle=black,
    colbacktitle=gray!20,
    attach boxed title to top left={xshift=6pt, yshift=-2pt},
    boxed title style={colframe=gray!50, sharp corners, boxrule=0.3pt},
    breakable
]
Is this picture strange? Answer with ``Yes'' if there is anything strange or abnormal in the image, such as more than 2 arms, more than 2 legs, more than 2 faces, missing body parts, or any glaring issue with the generated person. If you do not see any person in the image, or if the person is very small and hard to see, consider it an abnormality as well. Answer with ``No'' if there is nothing strange or abnormal in the image.
\end{tcolorbox}

\subsection{MLLM geolocalization prompt}\label{prompt:mllm_geoloc}
For all MLLMs, we use the same deterministic prompt requiring a single latitude--longitude prediction:

\begin{tcolorbox}[
    enhanced,
    colback=gray!5,
    colframe=gray!50,
    boxrule=0.4pt,
    arc=2pt,
    left=8pt, right=8pt, top=6pt, bottom=6pt,
    fontupper=\itshape\small,
    title=\textbf{MLLM Geolocalization Prompt},
    coltitle=black,
    colbacktitle=gray!20,
    attach boxed title to top left={xshift=6pt, yshift=-2pt},
    boxed title style={colframe=gray!50, sharp corners, boxrule=0.3pt},
    breakable
]
Estimate the geographic location of this image. Use visual clues such as architecture, vegetation, terrain, road markings, signs, weather, vehicles, shadows, coastlines, mountains, water, and text. Return exactly one JSON object with numeric ``lat'' and ``lon'' values and nothing else. Use signed decimal degrees: south and west are negative, north and east are positive. Do not copy coordinates from examples or instructions; there are no reference coordinates in this prompt. Rules: no explanation, no markdown, no prose, no reasoning, no \texttt{<think>} section. If uncertain, still give your best guess as latitude and longitude. The first character of your answer must be ``\{'' and the last character must be ``\}''.
\end{tcolorbox}

We parse the returned JSON object and discard outputs that cannot be converted into valid numeric latitude and longitude values. This fixed-prompt protocol ensures that differences across MLLMs are attributable to model behavior rather than to prompt variation.

\section{Taxonomy}
\label{app:taxonomy}

We define the demographic taxonomy used in our prompt construction by mapping each broad group label to a set of associated countries. This taxonomy is used only as an operational grouping for experimental prompt generation and should not be interpreted as an exhaustive, fixed, or essentialized definition of cultural, ethnic, or national identity.

\begin{itemize}
    \item \textbf{Indian.} India, Pakistan, Bangladesh, Nepal, and Sri Lanka.

    \item \textbf{East Asian.} China, Japan, South Korea, North Korea, and Taiwan.

    \item \textbf{Southeast Asian.} Thailand, Vietnam, Indonesia, Malaysia, Philippines, Cambodia, Laos, and Myanmar.

    \item \textbf{Middle Eastern.} Saudi Arabia, Iran, Iraq, Israel, Jordan, United Arab Emirates, Oman, Yemen, Qatar, Kuwait, Syria, and Lebanon.

    \item \textbf{Black.} Algeria, Angola, Benin, Botswana, Burkina Faso, Burundi, Cameroon, Cabo Verde, Central African Republic, Chad, Comoros, Democratic Republic of the Congo, Congo, Côte d'Ivoire, Djibouti, Egypt, Equatorial Guinea, Eritrea, Eswatini, Ethiopia, Gabon, Gambia, Ghana, Guinea, Guinea-Bissau, Kenya, Lesotho, Liberia, Libya, Madagascar, Malawi, Mali, Mauritania, Mauritius, Morocco, Mozambique, Namibia, Niger, Nigeria, Rwanda, São Tomé and Príncipe, Senegal, Seychelles, Sierra Leone, Somalia, South Africa, South Sudan, Sudan, Tanzania, Togo, Tunisia, Uganda, Zambia, and Zimbabwe.

    \item \textbf{Latino/Hispanic.} Mexico, Guatemala, Belize, Honduras, El Salvador, Nicaragua, Costa Rica, Panama, Cuba, Dominican Republic, Haiti, Jamaica, Trinidad and Tobago, Colombia, Venezuela, Guyana, Suriname, Ecuador, Peru, Brazil, Bolivia, Paraguay, Uruguay, Argentina, and Chile.

    \item \textbf{White.} United States, Canada, United Kingdom, Ireland, France, Germany, Italy, Spain, Portugal, Netherlands, Belgium, Switzerland, Austria, Greece, Poland, Sweden, Norway, Finland, Denmark, Australia, New Zealand, Russia, and Ukraine.
\end{itemize}

\section{Additional Results}
\subsection{Additional Results on the Effect of Race-Associated Attire}~\label{suppl:results_attire}
We report the effect of race-associated attire on the other models in Tabs.~\ref{tab:img2loc-attire-heatmap}, \ref{tab:g3-attire-heatmap}, \ref{tab:deepseek-attire-heatmap}, \ref{tab:model-attire-heatmap_intern}, \ref{tab:qwen3vl-attire-heatmap}, and \ref{tab:glm41v-attire-heatmap}.

\begin{table*}[t]
\centering
\caption{
\textbf{Effect of explicit race-associated attire conditioning across distance thresholds.}
Each subtable reports accuracy in the race-associated condition minus accuracy in the attire-unconditioned condition, in percentage points. Negative values indicate lower accuracy under race-associated explicit attire conditioning. The attire-unconditioned setting omits clothing descriptions but can retain the generator's demographic clothing priors.
}
\label{tab:attire-heatmap-grid}

\begin{subtable}[t]{0.48\textwidth}
\centering
\caption{Img2Loc}
\resizebox{\linewidth}{!}{%
\begin{tabular}{lrrrrr}
\hline
Race & 1km & 25km & 200km & 750km & 2500km \\ \hline
Black            & \hA{0.00}  & \hB{-1.6} & \hC{-4.8} & \hE{-9.9} & \hF{-11.7} \\
East Asian       & \hA{-0.2} & \hA{-0.6} & \hB{-2.5} & \hD{-6.1} & \hE{-8.6}  \\
Indian           & \hA{1.0}  & \hA{1.2}  & \hA{0.6}  & \hA{0.4}  & \hB{-3.2}  \\
Latino           & \hA{-0.6} & \hB{-2.0} & \hB{-2.5} & \hC{-4.7} & \hC{-4.7}  \\
Middle Eastern   & \hA{0.9}  & \hA{-0.3} & \hA{0.3}  & \hA{0.6}  & \hA{0.7}   \\
Southeast Asian  & \hA{0.1}  & \hA{0.6}  & \hA{-1.5} & \hB{-2.3} & \hB{-2.3}  \\
White            & \hA{-0.3} & \hA{-0.2} & \hA{-0.7} & \hA{-0.2} & \hA{1.3}   \\ \hline
\end{tabular}%
}
\label{tab:img2loc-attire-heatmap}
\end{subtable}
\hfill
\begin{subtable}[t]{0.48\textwidth}
\centering
\caption{G3}
\resizebox{\linewidth}{!}{%
\begin{tabular}{lrrrrr}
\hline
Race & 1km & 25km & 200km & 750km & 2500km \\ \hline
Black            & \hB{-1.7} & \hD{-6.1} & \hE{-8.8} & \hF{-13.0} & \hF{-17.5} \\
East Asian       & \hA{0.4}  & \hB{-2.8} & \hC{-3.5} & \hE{-8.2}  & \hE{-8.5}  \\
Indian           & \hA{-0.6} & \hA{0.3}  & \hB{-2.0} & \hC{-4.5}  & \hD{-5.8}  \\
Latino Hispanic  & \hA{-1.0} & \hC{-3.4} & \hC{-4.5} & \hD{-6.3}  & \hC{-4.8}  \\
Middle Eastern   & \hA{-0.7} & \hB{-2.5} & \hB{-1.8} & \hB{-2.0}  & \hA{-1.0}  \\
Southeast Asian  & \hA{-1.0} & \hC{-3.9} & \hC{-3.6} & \hD{-6.4}  & \hE{-8.3}  \\
White            & \hB{-1.5} & \hC{-3.5} & \hC{-3.5} & \hA{-0.4}  & \hB{-2.6}  \\ \hline
\end{tabular}%
}
\label{tab:g3-attire-heatmap}
\end{subtable}

\vspace{1em}

\begin{subtable}[t]{0.48\textwidth}
\centering
\caption{DeepSeek-VL}
\resizebox{\linewidth}{!}{%
\begin{tabular}{lrrrrr}
\hline
Race & 1km & 25km & 200km & 750km & 2500km \\ \hline
Black            & \hA{0.0}  & \hA{-1.3} & \hB{-3.2} & \hD{-5.1} & \hE{-8.0} \\
East Asian       & \hA{-0.1} & \hA{-1.6} & \hB{-2.8} & \hC{-3.8} & \hC{-5.0} \\
Indian           & \hA{0.0}  & \hA{-0.9} & \hB{-2.5} & \hB{-3.1} & \hC{-4.4} \\
Latino Hispanic  & \hA{-0.1} & \hA{-1.3} & \hB{-3.2} & \hC{-3.4} & \hD{-6.3} \\
Middle Eastern   & \hA{0.0}  & \hA{-0.3} & \hA{0.3}  & \hB{-1.8} & \hB{-3.2} \\
Southeast Asian  & \hA{0.0}  & \hA{-0.9} & \hB{-3.2} & \hC{-3.8} & \hB{-2.8} \\
White            & \hA{0.1}  & \hA{0.0}  & \hA{-0.9} & \hA{-0.1} & \hA{-1.5} \\ \hline
\end{tabular}%
}
\label{tab:deepseek-attire-heatmap}
\end{subtable}
\hfill
\begin{subtable}[t]{0.48\textwidth}
\centering
\caption{InternVL3.5}
\resizebox{\linewidth}{!}{%
\begin{tabular}{lrrrrr}
\hline
Race & 1km & 25km & 200km & 750km & 2500km \\ \hline
White            & \hA{0.0}  & \hA{-0.6} & \hA{0.7}  & \hB{-3.1} & \hB{-3.1} \\
Black            & \hA{-0.2} & \hA{-0.6} & \hB{-2.6} & \hC{-5.1} & \hE{-8.0} \\
East Asian       & \hA{0.1}  & \hA{-1.3} & \hA{-1.5} & \hC{-5.2} & \hD{-7.4} \\
Indian           & \hA{-0.3} & \hA{0.0}  & \hA{-0.7} & \hB{-2.6} & \hA{-1.3} \\
Latino           & \hA{-0.2} & \hA{-0.9} & \hA{-1.5} & \hB{-2.0} & \hA{-1.9} \\
Middle Eastern   & \hA{-0.3} & \hA{-0.7} & \hA{-1.2} & \hC{-3.5} & \hB{-2.2} \\
Southeast Asian  & \hA{-0.3} & \hA{-0.3} & \hA{-1.2} & \hB{-2.5} & \hA{0.3}  \\ \hline
\end{tabular}%
}
\label{tab:model-attire-heatmap_intern}
\end{subtable}

\vspace{1em}

\begin{subtable}[t]{0.48\textwidth}
\centering
\caption{Qwen3-VL}
\resizebox{\linewidth}{!}{%
\begin{tabular}{lrrrrr}
\hline
Race & 1km & 25km & 200km & 750km & 2500km \\ \hline
Black            & \hA{0.3}  & \hB{-1.8} & \hC{-4.5} & \hD{-6.1}  & \hE{-8.6}  \\
East Asian       & \hA{0.0}  & \hA{-1.3} & \hC{-4.5} & \hD{-6.6}  & \hD{-7.3}  \\
Indian           & \hA{-0.4} & \hA{-0.6} & \hB{-1.8} & \hD{-5.7}  & \hC{-4.5}  \\
Latino Hispanic  & \hA{0.3}  & \hB{-2.2} & \hA{-1.5} & \hD{-6.4}  & \hD{-5.7}  \\
Middle Eastern   & \hA{0.0}  & \hB{-1.6} & \hB{-2.6} & \hC{-3.8}  & \hA{-0.4}  \\
Southeast Asian  & \hA{-0.3} & \hC{-3.4} & \hD{-5.7} & \hF{-11.5} & \hF{-13.7} \\
White            & \hA{-0.9} & \hA{0.2}  & \hA{0.2}  & \hA{0.6}   & \hA{-0.3}  \\ \hline
\end{tabular}%
}
\label{tab:qwen3vl-attire-heatmap}
\end{subtable}
\hfill
\begin{subtable}[t]{0.48\textwidth}
\centering
\caption{GLM-4.1V}
\resizebox{\linewidth}{!}{%
\begin{tabular}{lrrrrr}
\hline
Race & 1km & 25km & 200km & 750km & 2500km \\ \hline
Black            & \hA{0.3}  & \hB{-2.9} & \hD{-7.6} & \hD{-7.1}  & \hE{-8.5} \\
East Asian       & \hA{0.0}  & \hC{-4.4} & \hD{-7.9} & \hF{-10.4} & \hE{-8.3} \\
Indian           & \hA{-0.1} & \hA{-0.7} & \hB{-2.6} & \hC{-5.1}  & \hB{-3.7} \\
Latino Hispanic  & \hA{-0.4} & \hB{-3.2} & \hC{-4.2} & \hD{-6.7}  & \hD{-7.2} \\
Middle Eastern   & \hA{-1.0} & \hA{-0.7} & \hA{-0.6} & \hB{-2.2}  & \hA{0.2}  \\
Southeast Asian  & \hA{-0.2} & \hB{-3.1} & \hD{-7.3} & \hE{-8.9}  & \hE{-9.8} \\
White            & \hA{0.3}  & \hA{0.3}  & \hA{-0.6} & \hB{-2.2}  & \hA{-1.0} \\ \hline
\end{tabular}%
}
\label{tab:glm41v-attire-heatmap}
\end{subtable}

\end{table*}

\subsection{Additional Results on Landmark Sensitivity}\label{suppl:results_landmark}
\begin{figure*}
    \centering

    \begin{subfigure}[t]{0.48\textwidth}
        \centering
        \begin{tikzpicture}
        \begin{axis}[
            ybar, width=\linewidth, height=0.78\linewidth, bar width=15pt,
            enlarge x limits=0.48, ylabel={\%}, ylabel style={font=\small},
            symbolic x coords={POPI-H, POPI-D}, xtick=data,
            xticklabels={POPI-H, POPI-D},
            tick label style={font=\small}, x tick label style={font=\small},
            yticklabel style={font=\small},
            ymin=0, ymax=50, ytick={0,10,20,30,40},
            ymajorgrids=true, grid style={dashed, gray!30},
            axis line style={black!70},
            nodes near coords,
            nodes near coords style={font=\scriptsize, /pgf/number format/fixed, /pgf/number format/precision=1},
            every node near coord/.append style={yshift=1.5pt},
            legend style={at={(0.5,1.08)}, anchor=south, legend columns=2, font=\scriptsize, draw=none, fill=none, column sep=4pt},
            legend image code/.code={\draw[#1, draw=black] (0cm,-0.08cm) rectangle (0.16cm,0.08cm);}
        ]
        \addplot[fill=ibmBlue, draw=black]   coordinates {(POPI-H, 14.58) (POPI-D, 30.49)};
        \addplot[fill=ibmOrange, draw=black] coordinates {(POPI-H, 29.60) (POPI-D, 39.22)};
        \legend{Landmark, Non-landmark}
        \end{axis}
        \end{tikzpicture}
        \caption{Img2Loc}
        \label{fig:img2loc-popi}
    \end{subfigure}
    \hfill
    \begin{subfigure}[t]{0.48\textwidth}
        \centering
        \begin{tikzpicture}
        \begin{axis}[
            ybar, width=\linewidth, height=0.78\linewidth, bar width=15pt,
            enlarge x limits=0.48, ylabel={\%}, ylabel style={font=\small},
            symbolic x coords={POPI-H, POPI-D}, xtick=data,
            xticklabels={POPI-H, POPI-D},
            tick label style={font=\small}, x tick label style={font=\small},
            yticklabel style={font=\small},
            ymin=0, ymax=50, ytick={0,10,20,30,40},
            ymajorgrids=true, grid style={dashed, gray!30},
            axis line style={black!70},
            nodes near coords,
            nodes near coords style={font=\scriptsize, /pgf/number format/fixed, /pgf/number format/precision=1},
            every node near coord/.append style={yshift=1.5pt},
            legend style={at={(0.5,1.08)}, anchor=south, legend columns=2, font=\scriptsize, draw=none, fill=none, column sep=4pt},
            legend image code/.code={\draw[#1, draw=black] (0cm,-0.08cm) rectangle (0.16cm,0.08cm);}
        ]
        \addplot[fill=ibmBlue, draw=black]   coordinates {(POPI-H, 13.54) (POPI-D, 26.46)};
        \addplot[fill=ibmOrange, draw=black] coordinates {(POPI-H, 23.54) (POPI-D, 31.02)};
        \legend{Landmark, Non-landmark}
        \end{axis}
        \end{tikzpicture}
        \caption{G3}
        \label{fig:g3-popi}
    \end{subfigure}

    \vspace{1em}

    \begin{subfigure}[t]{0.48\textwidth}
        \centering
        \begin{tikzpicture}
        \begin{axis}[
            ybar, width=\linewidth, height=0.78\linewidth, bar width=15pt,
            enlarge x limits=0.48, ylabel={\%}, ylabel style={font=\small},
            symbolic x coords={POPI-H, POPI-D}, xtick=data,
            xticklabels={POPI-H, POPI-D},
            tick label style={font=\small}, x tick label style={font=\small},
            yticklabel style={font=\small},
            ymin=0, ymax=50, ytick={0,10,20,30,40},
            ymajorgrids=true, grid style={dashed, gray!30},
            axis line style={black!70},
            nodes near coords,
            nodes near coords style={font=\scriptsize, /pgf/number format/fixed, /pgf/number format/precision=1},
            every node near coord/.append style={yshift=1.5pt},
            legend style={at={(0.5,1.08)}, anchor=south, legend columns=2, font=\scriptsize, draw=none, fill=none, column sep=4pt},
            legend image code/.code={\draw[#1, draw=black] (0cm,-0.08cm) rectangle (0.16cm,0.08cm);}
        ]
        \addplot[fill=ibmBlue, draw=black]   coordinates {(POPI-H, 12.71) (POPI-D, 14.10)};
        \addplot[fill=ibmOrange, draw=black] coordinates {(POPI-H, 22.87) (POPI-D, 20.12)};
        \legend{Landmark, Non-landmark}
        \end{axis}
        \end{tikzpicture}
        \caption{DeepSeek-VL}
        \label{fig:deepseekvl-popi}
    \end{subfigure}
    \hfill
    \begin{subfigure}[t]{0.48\textwidth}
        \centering
        \begin{tikzpicture}
        \begin{axis}[
            ybar, width=\linewidth, height=0.78\linewidth, bar width=15pt,
            enlarge x limits=0.48, ylabel={\%}, ylabel style={font=\small},
            symbolic x coords={POPI-H, POPI-D}, xtick=data,
            xticklabels={POPI-H, POPI-D},
            tick label style={font=\small}, x tick label style={font=\small},
            yticklabel style={font=\small},
            ymin=0, ymax=50, ytick={0,10,20,30,40},
            ymajorgrids=true, grid style={dashed, gray!30},
            axis line style={black!70},
            nodes near coords,
            nodes near coords style={font=\scriptsize, /pgf/number format/fixed, /pgf/number format/precision=1},
            every node near coord/.append style={yshift=1.5pt},
            legend style={at={(0.5,1.08)}, anchor=south, legend columns=2, font=\scriptsize, draw=none, fill=none, column sep=4pt},
            legend image code/.code={\draw[#1, draw=black] (0cm,-0.08cm) rectangle (0.16cm,0.08cm);}
        ]
        \addplot[fill=ibmBlue, draw=black]   coordinates {(POPI-H, 16.67) (POPI-D, 19.03)};
        \addplot[fill=ibmOrange, draw=black] coordinates {(POPI-H, 28.81) (POPI-D, 26.76)};
        \legend{Landmark, Non-landmark}
        \end{axis}
        \end{tikzpicture}
        \caption{InternVL3.5}
        \label{fig:intern-popi}
    \end{subfigure}

    \vspace{1em}

    \begin{subfigure}[t]{0.48\textwidth}
        \centering
        \begin{tikzpicture}
        \begin{axis}[
            ybar, width=\linewidth, height=0.78\linewidth, bar width=15pt,
            enlarge x limits=0.48, ylabel={\%}, ylabel style={font=\small},
            symbolic x coords={POPI-H, POPI-D}, xtick=data,
            xticklabels={POPI-H, POPI-D},
            tick label style={font=\small}, x tick label style={font=\small},
            yticklabel style={font=\small},
            ymin=0, ymax=50, ytick={0,10,20,30,40},
            ymajorgrids=true, grid style={dashed, gray!30},
            axis line style={black!70},
            nodes near coords,
            nodes near coords style={font=\scriptsize, /pgf/number format/fixed, /pgf/number format/precision=1},
            every node near coord/.append style={yshift=1.5pt},
            legend style={at={(0.5,1.08)}, anchor=south, legend columns=2, font=\scriptsize, draw=none, fill=none, column sep=4pt},
            legend image code/.code={\draw[#1, draw=black] (0cm,-0.08cm) rectangle (0.16cm,0.08cm);}
        ]
        \addplot[fill=ibmBlue, draw=black]   coordinates {(POPI-H, 7.08) (POPI-D, 9.72)};
        \addplot[fill=ibmOrange, draw=black] coordinates {(POPI-H, 23.21) (POPI-D, 22.22)};
        \legend{Landmark, Non-landmark}
        \end{axis}
        \end{tikzpicture}
        \caption{Qwen3-VL}
        \label{fig:qwen-popi}
    \end{subfigure}
    \hfill
    \begin{subfigure}[t]{0.48\textwidth}
        \centering
        \begin{tikzpicture}
        \begin{axis}[
            ybar, width=\linewidth, height=0.78\linewidth, bar width=15pt,
            enlarge x limits=0.48, ylabel={\%}, ylabel style={font=\small},
            symbolic x coords={POPI-H, POPI-D}, xtick=data,
            xticklabels={POPI-H, POPI-D},
            tick label style={font=\small}, x tick label style={font=\small},
            yticklabel style={font=\small},
            ymin=0, ymax=50, ytick={0,10,20,30,40},
            ymajorgrids=true, grid style={dashed, gray!30},
            axis line style={black!70},
            nodes near coords,
            nodes near coords style={font=\scriptsize, /pgf/number format/fixed, /pgf/number format/precision=1},
            every node near coord/.append style={yshift=1.5pt},
            legend style={at={(0.5,1.08)}, anchor=south, legend columns=2, font=\scriptsize, draw=none, fill=none, column sep=4pt},
            legend image code/.code={\draw[#1, draw=black] (0cm,-0.08cm) rectangle (0.16cm,0.08cm);}
        ]
        \addplot[fill=ibmBlue, draw=black]   coordinates {(POPI-H, 4.1) (POPI-D, 7.64)};
        \addplot[fill=ibmOrange, draw=black] coordinates {(POPI-H, 20.52) (POPI-D, 21.1)};
        \legend{Landmark, Non-landmark}
        \end{axis}
        \end{tikzpicture}
        \caption{GLM-4.1V}
        \label{fig:glm-popi}
    \end{subfigure}

    \caption{
    \textbf{Landmark sensitivity analysis across six models.}
    POPI-H and POPI-D for landmark and non-landmark images. Non-landmark images exhibit larger prediction shifts across all models, suggesting that the inserted person has a stronger effect when distinctive geographic landmarks are absent.
    }
    \label{fig:popi-grid}
\end{figure*}
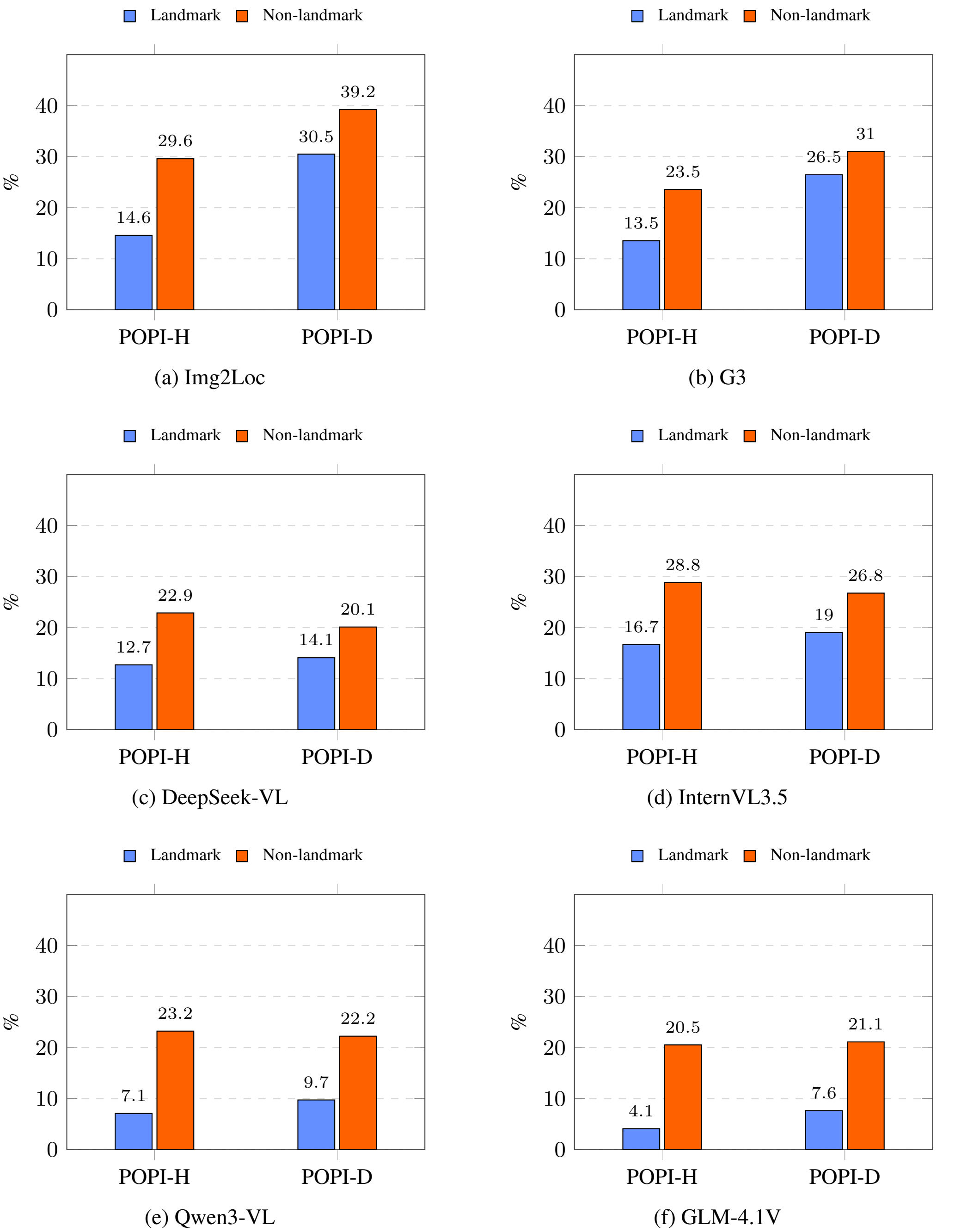

In Figs.~\ref{fig:img2loc-popi}, \ref{fig:g3-popi}, \ref{fig:deepseekvl-popi}, \ref{fig:intern-popi}, \ref{fig:qwen-popi}, and \ref{fig:glm-popi} we report the POPI-H and POPI-D values for landmark and non-landmark images for all the other models.

\subsection{Additional Results on POPI}
To complement the aggregate results, we evaluate POPI metrics separately across each inserted demographic group (Tab.~\ref{tab:race_popi}). We observe substantial variation across categories: for GeoCLIP, Black subjects yield the highest directional steering ($\text{POPI-A} = 37.1\%$) and distraction ($\text{POPI-D} = 38.4\%$), while Middle Eastern subjects produce the highest source-region correction rate ($\text{POPI-H} = 38.3\%$). Conversely, White subjects exhibit the lowest values across all three indicators ($\le 17.6\%$).

These group-level differences capture model reliance on learned demographic shortcuts under our operational support regions rather than intrinsic properties of the depicted groups. Because transition rates depend on eligible baseline predictions, visual subject cues, and region boundaries, variations across groups reflect model-specific geographic associations. Finally, we reiterate that POPI-H quantifies incidental source-region transitions caused by an irrelevant foreground intervention and does not represent an improvement in model fairness.

\begin{table}[t]
\centering
\caption{\textbf{GeoCLIP POPI metrics by inserted demographic group.}
Scores are computed over the corresponding eligible baseline populations defined in Sec.~\ref{sec:popi}. Values are percentages.}
\label{tab:race_popi}
\small
\setlength{\tabcolsep}{8pt}
\begin{tabular}{lrrr}
\toprule
Inserted Group & POPI-A & POPI-H & POPI-D \\
\midrule
Black           & 37.1 & 29.3 & 38.4 \\
East Asian      & 28.7 & 17.0 & 30.7 \\
Indian          & 30.0 & 15.4 & 32.5 \\
Latino Hispanic         & 30.2 & 16.0 & 32.6 \\
Middle Eastern  & 29.5 & 38.3 & 28.0 \\
Southeast Asian & 33.0 & 19.7 & 35.2 \\
White           & 16.5 & 10.1 & 17.6 \\
\bottomrule
\end{tabular}
\end{table}

\vspace{1em}

\subsection{Additional Results on Larger MLLMs}
\label{app:large_mllms}
To assess whether demographic sensitivity persists in higher-capacity architectures, we extend our evaluation to InternVL3.5-38B\cite{wang2025internvl3}, Gemma3-27B-IT\cite{gemma_2025}, and Qwen3VL-30B-A3B\cite{qwen3technicalreport}. As reported in Tab.~\ref{tab:large_mllms}, all three models exhibit systematic accuracy drops following foreground person insertion across all evaluated geodesic thresholds. At the 2500\,km continent level, accuracy decreases by 15.7 percentage points for InternVL3.5-38B (from 71.1\% to 55.4\%), 12.8 points for Gemma3-27B-IT (from 80.6\% to 67.8\%), and 13.4 points for Qwen3VL-30B-A3B (from 86.3\% to 72.9\%).

Furthermore, non-zero POPI rates confirm that these errors are directional rather than unstructured degradations, demonstrating that susceptibility to foreground demographic cues remains present in larger MLLMs. Because these models also differ in pretraining mixtures, vision-backbone resolution, and instruction tuning, this comparison indicates persistence across scale rather than isolating parameter count as a single controlled variable.

\begin{table}[t]
\centering
\caption{\textbf{Geolocalization accuracy and POPI metrics for larger MLLMs.} Accuracy at standard geodesic distance thresholds and POPI transition rates are reported as percentages. Original and Edited denote unmodified source backgrounds and their counterfactual edited variants, respectively.}
\label{tab:large_mllms}
\scriptsize
\setlength{\tabcolsep}{3.5pt}
\begin{tabular}{ccccccl|ccc}
\hline
Methods &
  Edited &
  \begin{tabular}[c]{@{}c@{}}Street\\ 1km\end{tabular} &
  \begin{tabular}[c]{@{}c@{}}City\\ 25km\end{tabular} &
  \begin{tabular}[c]{@{}c@{}}Region\\ 200km\end{tabular} &
  \begin{tabular}[c]{@{}c@{}}Country\\ 750km\end{tabular} &
  \begin{tabular}[l]{@{}c@{}}Continent\\ 2500km\end{tabular} &
  POPI-A ($\downarrow$) &
  POPI-H ($\downarrow$) &
  POPI-D ($\downarrow$)\\ \hline

\multirow{2}{*}{InternVL3.5-38B} &
  \xmark &
  1.2 &
  15.2 &
  28.6 &
  48.9 &
  71.1 &
  \multirow{2}{*}{23.2} &
  \multirow{2}{*}{21.4} &
  \multirow{2}{*}{23.4} \\
 &
  \cmark &
  1.0\drop{17} &
  11.0\drop{28} &
  19.5\drop{32} &
  36.5\drop{25} &
  55.4\otherdrop{22} &
  & & \\
\hline

\multirow{2}{*}{Gemma3-27B-IT} &
  \xmark &
  5.5 &
  30.7 &
  48.0 &
  68.7 &
  80.6 &
  \multirow{2}{*}{15.7} &
  \multirow{2}{*}{19.2} &
  \multirow{2}{*}{15.1} \\
 &
  \cmark &
  3.8\drop{31} &
  24.6\drop{20} &
  37.8\drop{21} &
  54.4\drop{21} &
  67.8\otherdrop{16} &
  & & \\
\hline

\multirow{2}{*}{Qwen3VL-30B-A3B} &
  \xmark &
  5.2 &
  33.7 &
  50.2 &
  74.5 &
  86.3 &
  \multirow{2}{*}{14.3} &
  \multirow{2}{*}{14.4} &
  \multirow{2}{*}{14.3} \\
 &
  \cmark &
  4.4\drop{15} &
  29.3\drop{13} &
  41.8\drop{17} &
  60.4\drop{19} &
  72.9\otherdrop{16} &
  & & \\
\hline

\end{tabular}
\end{table}

\section{Computational Resources}\label{app:computational_resources}
In this section, we provide additional details regarding the computational resources required to replicate the experiments reported in this paper.
The image generation pipeline was run on an NVIDIA RTX 5000 Blackwell and it took approximately 10 hours. Model inference was done on an NVIDIA H100 and it took no more than one day to run all experiments.
\section{Dataset Licences}\label{app:licences}
In this section, we mention the licenses under which each of the datasets we cite are released:
\begin{itemize}
    \item Im2GPS3k: Apache 2.0
    \item YFCC4k: CC BY-NC-SA
    \item FairFace: CC BY 4.0
    \item \textsc{GeoBiaset}: CC BY-NC 4.0
\end{itemize}
\newpage
\section{Foreground Masking Controls}
\label{app:causal_isolation}
The comparison between original backgrounds and their counterfactual edits does not by itself separate three distinct factors: (i) geographic evidence lost through foreground occlusion, (ii) background drift introduced by the generative editing pipeline, and (iii) demographic cues carried by the inserted subject. We therefore perform a paired masking experiment to isolate the first two factors.

Using SAM3~\cite{carion2025sam3segmentconcepts}, we segment the inserted subject in each of the 9,212 edited images. Each variant receives an instance-specific mask capturing the exact shape, scale, and placement of the generated person (using the union of masks if multiple instances are generated). Crucially, each mask is applied identically to both the edited image and its matched original background, ensuring identical occlusions so that residual differences reflect solely whether the visible background is authentic or synthesized.
We evaluate three masking fills:
(i) constant black,
(ii) uniform gray ($\text{RGB} = [128, 128, 128]$), and
(iii) Gaussian blur ($\sigma = 15$ pixels).
Black and gray eliminate all visual features in the subject region, whereas Gaussian blur preserves coarse local illumination and chromaticity.

\begin{table}[t]
\centering
\caption{
\textbf{Geolocalization accuracy for the foreground-masking control experiment.}
For each edited image, the instance-specific SAM3 mask of the inserted subject is applied identically to its matched original background. Accuracies are reported as percentages varying the distance thresholds.
}
\label{tab:sam3_masking}
\small
\setlength{\tabcolsep}{5pt}
\begin{tabular}{llccccc}
\toprule
Source & Mask & $\le$1\,km & $\le$25\,km & $\le$200\,km & $\le$750\,km & $\le$2500\,km \\
\midrule
Original & None  & 11.1 & 35.6 & 51.6 & 68.8 & 81.9 \\
Edited   & None  &  4.3 & 16.3 & 25.4 & 38.1 & 55.3 \\
\midrule
Original & Black &  8.1 & 28.4 & 40.9 & 57.9 & 73.4 \\
Edited   & Black &  8.1 & 27.8 & 40.2 & 56.1 & 72.7 \\
\midrule
Original & Gray  &  7.5 & 27.7 & 40.8 & 58.6 & 73.6 \\
Edited   & Gray  &  7.4 & 26.7 & 39.5 & 56.0 & 71.7 \\
\midrule
Original & Blur  &  8.9 & 29.4 & 42.7 & 60.6 & 76.7 \\
Edited   & Blur  &  7.8 & 24.8 & 36.8 & 53.6 & 70.9 \\
\bottomrule
\end{tabular}
\end{table}

\minisection{Effect of foreground occlusion.}
Occluding the subject region on original backgrounds reduces continent-level accuracy from $81.9\%$ to $73.4\%$ (black), $73.6\%$ (gray), and $76.7\%$ (blur), confirming that physical occlusion removes useful scene evidence. However, this drop is far smaller than the degradation observed when the synthetic person is visible ($55.3\%$). Occlusion alone therefore cannot explain the observed performance loss.

\minisection{Effect of generation-induced background changes.}
Comparing masked originals to masked edits isolates distortions outside the subject mask. Under black masking, continent-level accuracy differs by only $0.7$ points ($73.4\%$ vs.\ $72.7\%$), and under gray masking by $1.9$ points ($73.6\%$ vs.\ $71.7\%$). Although full-reference metrics (Sec.~\ref{sec:quality}) confirm that pixel-level changes occur, these controls demonstrate that masked generative backgrounds retain nearly all task-relevant geographic evidence used by the model.

\minisection{Approximate decomposition of continent-level degradation.}
For GeoCLIP, the total continent-level degradation between original and edited images is $26.6$ percentage points ($81.9\%$ vs.\ $55.3\%$). Combining the black-mask control with the face-obfuscation results 
(Sec.~\ref{sec:results_anonymized}) yields an approximate loss attribution:
\begin{itemize}
    \item \textbf{Scene occlusion:} $8.5$ points ($31.9\%$ of the drop), measured by masking the original image.
    \item \textbf{Background drift:} $0.7$ points ($2.6\%$), measured by swapping the masked original with the masked edit.
    \item \textbf{Non-facial subject cues:} $9.0$ points ($33.9\%$), revealed by inserting the subject with an obfuscated face.
    \item \textbf{Facial appearance:} $8.4$ points ($31.6\%$), measured when blurring the face.
\end{itemize}
While diagnostic rather than a formal causal factorization, this shows that approximately $65.5\%$ of the degradation stems directly from demographic features of the inserted person (facial and non-facial) rather than from scene destruction or editing artifacts.

\clearpage

\end{appendices}

\newpage
\section*{NeurIPS Paper Checklist}

\begin{enumerate}

\item {\bf Claims}
    \item[] Question: Do the main claims made in the abstract and introduction accurately reflect the paper's contributions and scope?
    \item[] Answer: \answerYes{} 
    \item[] Justification: The abstract and introduction state that the paper introduces \textsc{GeoBiaset}, a counterfactual benchmark for auditing demographic bias in worldwide image geolocalization, together with POPI metrics for measuring person-induced prediction shifts. These claims are supported by the dataset construction in Section \ref{sec:geobiaset}, the evaluation protocol in Section \ref{sec:protocol}, and the experimental results in Section \ref{sec:results}.

\item {\bf Limitations}
    \item[] Question: Does the paper discuss the limitations of the work performed by the authors?
    \item[] Answer: \answerYes{} 
    \item[] Justification: Section \ref{sec:ethics_limitations} discusses limitations including the simplified use of race and binary gender categories, the operational nature of race-associated geographic regions and clothing prompts, synthetic-to-real distribution shift, residual generation artifacts, and the limited focus on foreground people rather than other geolocalization cues.

\item {\bf Theory assumptions and proofs}
    \item[] Question: For each theoretical result, does the paper provide the full set of assumptions and a complete (and correct) proof?
    \item[] Answer: \answerNA{} 
    \item[] Justification: The paper does not introduce theoretical results, theorems, or proofs.

    \item {\bf Experimental result reproducibility}
    \item[] Question: Does the paper fully disclose all the information needed to reproduce the main experimental results of the paper to the extent that it affects the main claims and/or conclusions of the paper (regardless of whether the code and data are provided or not)?
    \item[] Answer: \answerYes{} 
    \item[] Justification: The paper describes the dataset construction procedure, image-editing pipeline, quality-control prompt, MLLM geolocalization prompt, evaluated models, geodesic accuracy metrics, and POPI definitions. These details provide a reasonable path to reproducing or verifying the main experimental claims.

\item {\bf Open access to data and code}
    \item[] Question: Does the paper provide open access to the data and code, with sufficient instructions to faithfully reproduce the main experimental results, as described in supplemental material?
    \item[] Answer: \answerYes{} 
    \item[] Justification: We release anonymized code at link \url{https://anonymous.4open.science/r/GeoBiaset/README.md} and the dataset at link \url{https://kaggle.com/datasets/6393b4d11746fe3e4af72d5c71438ea2e003283fec6044064becdfd82a68475a}

\item {\bf Experimental setting/details}
    \item[] Question: Does the paper specify all the training and test details (e.g., data splits, hyperparameters, how they were chosen, type of optimizer) necessary to understand the results?
    \item[] Answer: \answerYes{} 
    \item[] Justification: The paper specifies that the benchmark consists of original background images and counterfactual edited variants, describes the seven race categories, gender and attire conditions, evaluated geolocalization systems and MLLMs, distance thresholds, POPI metrics, and prompting details. Experimental setting is discussed in Sec. \ref{sec:geobiaset} Since the evaluated systems are used as existing models rather than newly trained models, optimizer and training hyperparameters are not applicable.

\item {\bf Experiment statistical significance}
    \item[] Question: Does the paper report error bars suitably and correctly defined or other appropriate information about the statistical significance of the experiments?
    \item[] Answer: \answerNo{} 
    \item[] Justification: We do not report confidence intervals or significance tests because all experiments are inference-only so run-to-run variance is negligible. Reported accuracy and POPI values are exact metrics computed over the full benchmark. 


\item {\bf Experiments compute resources}
    \item[] Question: For each experiment, does the paper provide sufficient information on the computer resources (type of compute workers, memory, time of execution) needed to reproduce the experiments?
    \item[] Answer: \answerYes{} 
    \item[] Justification: The paper discloses information about computational resources in App. \ref{app:computational_resources}.
    
\item {\bf Code of ethics}
    \item[] Question: Does the research conducted in the paper conform, in every respect, with the NeurIPS Code of Ethics \url{https://neurips.cc/public/EthicsGuidelines}?
    \item[] Answer: \answerYes{} 
    \item[] Justification: The paper explicitly frames \textsc{GeoBiaset} as an auditing benchmark for measuring demographic bias and discusses privacy, fairness, surveillance, and dual-use risks in Section \ref{sec:ethics_limitations}.

\item {\bf Broader impacts}
    \item[] Question: Does the paper discuss both potential positive societal impacts and negative societal impacts of the work performed?
    \item[] Answer: \answerYes{} 
    \item[] Justification: Section \ref{sec:ethics_limitations} discusses beneficial applications such as journalism, historical analysis, environmental monitoring, geography education, and open-source verification, while also acknowledging risks including surveillance, stalking, military targeting, privacy violations, and uneven harms from biased geolocation predictions.
    
\item {\bf Safeguards}
    \item[] Question: Does the paper describe safeguards that have been put in place for responsible release of data or models that have a high risk for misuse (e.g., pre-trained language models, image generators, or scraped datasets)?
    \item[] Answer: \answerYes{} 
    \item[] Justification: The dataset documentation and release materials describe intended use for bias evaluation and mitigation, prohibit harmful identification or surveillance use, and include warnings that race-associated regions and attire prompts are operational measurement constructs rather than normative mappings between people and places.


\item {\bf Licenses for existing assets}
    \item[] Question: Are the creators or original owners of assets (e.g., code, data, models), used in the paper, properly credited and are the license and terms of use explicitly mentioned and properly respected?
    \item[] Answer: \answerYes{} 
    \item[] Justification: The paper mentions the license of each dataset used in Appendix \ref{app:licences}.

\item {\bf New assets}
    \item[] Question: Are new assets introduced in the paper well documented and is the documentation provided alongside the assets?
    \item[] Answer: \answerYes{} 
    \item[] Justification: The paper introduces \textsc{GeoBiaset} as a new counterfactual benchmark and documents its construction in Section \ref{sec:geobiaset}, including the source background images, demographic taxonomy, face-pool selection, image-editing pipeline, quality-control procedure, and face-obfuscated variant. The appendices provide the exact image-editing, clothing, quality-control, and MLLM evaluation prompts, as well as the operational country-level taxonomy used for POPI evaluation.


\item {\bf Crowdsourcing and research with human subjects}
    \item[] Question: For crowdsourcing experiments and research with human subjects, does the paper include the full text of instructions given to participants and screenshots, if applicable, as well as details about compensation (if any)? 
    \item[] Answer: \answerNA{} 
    \item[] Justification: The paper does not involve crowdsourcing experiments.
\item {\bf Institutional review board (IRB) approvals or equivalent for research with human subjects}
    \item[] Question: Does the paper describe potential risks incurred by study participants, whether such risks were disclosed to the subjects, and whether Institutional Review Board (IRB) approvals (or an equivalent approval/review based on the requirements of your country or institution) were obtained?
    \item[] Answer: \answerNA{} 
    \item[] Justification: The paper does not involve research with human subjects.

\item {\bf Declaration of LLM usage}
    \item[] Question: Does the paper describe the usage of LLMs if it is an important, original, or non-standard component of the core methods in this research? Note that if the LLM is used only for writing, editing, or formatting purposes and does \emph{not} impact the core methodology, scientific rigor, or originality of the research, declaration is not required.
    \item[] Answer: \answerYes{} 
    \item[] Justification: The paper uses multimodal large language models as evaluated geolocalization systems and uses Qwen3-VL as an automated quality-control model for detecting severe synthesis failures in generated images. These uses are part of the experimental protocol and dataset construction pipeline, and the corresponding prompts are documented in Appendix \ref{app:prompts}.

\end{enumerate}

\end{document}